\def\paperlanguage{} %% English
\def\paperlanguage{}

\documentclass[letterpaper, 10 pt, conference]{ieeeconf}  % Comment this line out if you need a4paper

\usepackage{bm}
\usepackage{cite}
\usepackage{flushend}
\include{preamble}
\usepackage {booktabs}

\IEEEoverridecommandlockouts                              % This command is only needed if
\title{\LARGE \textbf
  {
    \switchlanguage%
    {%
      CubiX: Portable Wire-Driven Parallel Robot \\Connecting to and Utilizing the Environment
    }%
    {%
      CubiX: Portable Wire-Driven Parallel Robot \\Connecting to and Utilizing the Environment
    }%
  }
}

\author{Shintaro Inoue$^{1}$, Kento Kawaharazuka$^{1}$, Temma Suzuki$^{1}$, Sota Yuzaki$^{1}$, Kei Okada$^{1}$, and Masayuki Inaba$^{1}$% <-this % stops a space
  \thanks{$^{1}$ The authors are with the Department of Mechano-Informatics, Graduate School of Information Science and Technology, The University of Tokyo, 7-3-1 Hongo, Bunkyo-ku, Tokyo, 113-8656, Japan.
    {\texttt\small [s-inoue, kawaharazuka, t-suzuki, yuzaki, k-okada, inaba]@jsk.t.u-tokyo.ac.jp}
  }
}
\begin{document}

\maketitle
\thispagestyle{empty}
\pagestyle{empty}

%%%%%%%%%%%%%%%%%%%%%%%%%%%%%%%%%%%%%%%%%%%%%%%%%%%%%%%%%%%%%%%%%%%%%%%%%%%%%%%%
\begin{abstract}
  \switchlanguage%
  {%
    A wire-driven parallel robot is a type of robotic system 
    where multiple wires are used to control the movement of a end-effector. 
    The wires are attached to the end-effector and anchored to fixed points on external structures.
    This configuration allows for the separation of actuators and end-effectors,
    enabling lightweight and simplified movable parts in the robot.
    However, its range of motion remains confined within the space formed by the wires,
    limiting the wire-driven capability to only within the pre-designed operational range.
    Here, in this study, we develop a wire-driven robot, CubiX, capable of connecting to and utilizing the environment. % OK
    CubiX connects itself to the environment using up to 8 wires and drives itself by winding these wires. % OK
    By integrating actuators for winding the wires into CubiX,
    a portable wire-driven parallel robot is realized without limitations on its workspace. % OK
    Consequently, 
    the robot can form parallel wire-driven structures by connecting wires to the environment at any operational location. % OK
  }%
  {%
    ワイヤ駆動パラレルロボットは
    可動部であるエンドエフェクタに接続されたワイヤをアクチュエータにより巻き取ることで，
    ワイヤによって張られた空間内で駆動する．
    しかし，その空間はアクチュエータが固定された動作しないフレームの内側に留まっており，
    予め設計された活動範囲のみでしかワイヤ駆動による能力が発揮されない．
    そこで，本研究では，環境接続可能なワイヤ駆動ロボットCubiXを開発する．
    CubiXは自身と環境とを最大8本のワイヤで接続し，それらを自ら巻き取ることによって駆動する．
    ワイヤを巻き取るアクチュエータがCubiXに搭載されていることによって，
    活動場所が制限されないポータブルなワイヤ駆動パラレルロボットが実現した．
    これにより，ロボットが活動場所でワイヤを環境に接続して利用することで，パラレルワイヤ駆動を形成することができる．

    % 環境を用いることをもう少し強調しても良い
    % アクチュエータがエンドエフェクタから切り離されることにより，可動部の軽量化や単純化が実現する．
    % しかし，そのワイヤの配置はロボット体内に留まっており，% ロボット体内ってどこ？
    % アクチュエータを可動部のエンドエフェクタから切り離すことができる エンドエフェクタと言わずに，可動部でよくない？
    % ことにより，可動部の軽量化や単純化が実現する
    % しかし，可動部の可動範囲はアクチュエータが固定された内側のワイヤが張られる範囲に留まっており，
    % できることは変わらないが，1つのロボットに集約されているという構成法がいいのであるという主張になってなくないか TODO
  }%
\end{abstract}

\section{Introduction}\label{sec:introduction}
\switchlanguage%
{%
  Wire-driven robots, 
  capable of handling strong forces up to the wire breaking strength with high degrees of freedom provided by multiple wires, 
  have been developed extensively 
  \cite{8967836, 7759468, 525288, 6695742, cone1985skycam, xu2023collaborative, xu2023optimal,
  xu2024online, 5509299, merlet2010marionet, 8794265, 10375200}.

  Cable-driven parallel robots (CDPR) consist of a movable end effector, an external frame, and wires connecting them.
  The end effector is driven by actuators fixed to the external frame,
  which wind the wires connecting the frame and the end effector.
  The separation of actuators and end-effectors enables the lightweight and simplification of the movable part.
  In CoGiRo \cite{8967836}, a CDPR, 
  a box-shaped end effector with a robot arm is connected to the external frame via wires, 
  enabling the robot arm to move in 3-dimensional space by winding the wires.
  Similarly, CableRobot simulator \cite{7759468}, also a CDPR,
  is a rideable device capable of moving a human-occupied end effector in space at accelerations of up to 1.5 G,
  offering immersive experiences for driving simulations. 
  Such configurations, where the end effector is connected to the external frame via wires and driven through wire actuation, 
  are prevalent.
  Similar configurations include the ultra-high-speed wire-driven robot FALCON \cite{525288},
  capable of achieving accelerations of up to 43 G,
  as well as IPAnema \cite{6695742}, a CDPR developed for industrial use.
  Another example is SkyCam \cite{cone1985skycam}, deployed in sports stadiums to move cameras around the playing field.
  All these systems control the posture of the end effector using force transmission through multiple wires.
  However, the actuators generating the force
  % transmitted through the wires 
  are fixed to the external frame surrounding the end effector,
  limiting the robot's operational range within this frame.

  \begin{figure}[t]
    \begin{center}
      \includegraphics[width=1\columnwidth]{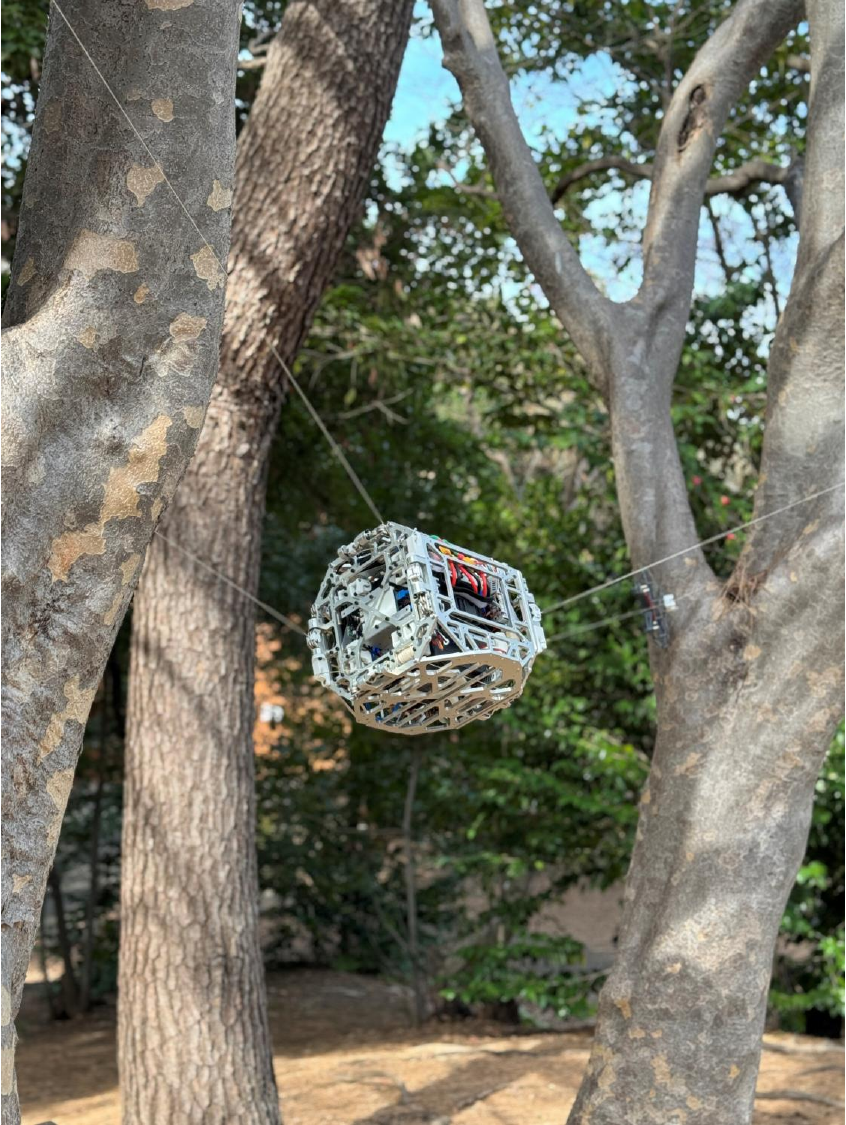}
      \vspace{-4.0ex}
      \caption{The overview of CubiX connecting to and utilizing the environment.}
      \vspace{-6.0ex}
      \label{fig:cubix}
    \end{center}
  \end{figure}

  In contrast, there are studies that enable the movement of the entire frame of CDPR 
  by employing multiple carts equipped with wire-winding modules \cite{xu2023collaborative,xu2023optimal,xu2024online}.
  However, as a result, 
  the robot's operational range is confined to the plane where the carts carrying the wire modules can move. 
  Marionet-CRANE \cite{5509299, merlet2010marionet}, another CDPR, consists of a series of cranes with wire modules attached, 
  requiring humans to install as many cranes as wires used in the operating area, 
  which does not necessarily free the CDPR from its external frame limitations.

  Turning our attention to robots that utilize the environment through wires,
  recent studies have explored improving vehicle mobility by connecting a wire from the vehicle to the environment via drones \cite{8794265},
  and enhancing humanoid capabilities through wire-connected carabiners, allowing the reinforcement 
  of forces exerted by the humanoid \cite{10375200}. 
  While wire-driven mechanisms in these studies were confined to linear force assistance, 
  utilizing the environment through wires is considered one effective option for wire-driven robots.

  Therefore, in this study, we develop CubiX (shown in \figref{fig:cubix}), 
  a wire-driven robot equipped with actuators capable of winding wires internally.
  CubiX connects multiple wires extending from its body to the environment and drives itself by winding these wires.
  Through this approach, 
  robots can form parallel wire-driven structures tailored to specific tasks by utilizing the environment at their operational sites.
}%
{%
  ワイヤ駆動ロボットは，ワイヤ破断強度までの強い力を
  複数のワイヤによる高い自由度で扱うことができ，
  これまでにも多く
  % のワイヤ駆動ロボットが
  開発されてきた
  \cite{8967836, 7759468, 525288, 6695742, cone1985skycam, xu2023collaborative, xu2023optimal, xu2024online, 
  5509299, merlet2010marionet, 8794265, 10375200}.

  % 説明　エンドエフェクタ　フレーム　定義　利点
  % ケーブル駆動パラレルロボット（以下，CDPR）であるCoGiRoは，
  ケーブル駆動パラレルロボット（以下，CDPR）は，
  可動部分であるエンドエフェクタと，それとは独立した構造部であるフレーム，それらを接続するワイヤからなる．
  フレームに固定されたアクチュエータがワイヤを巻き取ることによってエンドエフェクタが駆動する．
  エンドエフェクタからアクチュエータが切り離されることで，可動部の軽量化や単純化が実現される．
  CDPRであるCoGiRoは，ロボットアームがついた箱状のエンドエフェクタがワイヤによってフレームと接続され，
  ワイヤが巻き取られることによってロボットアームの3次元空間上の移動を実現した\cite{8967836}．
  同じくCDPRであるCableRobot simulatorは，搭乗型のCDPRであり，
  人が乗ったエンドエフェクタをワイヤ駆動によって最大1.5Gの加速度で空間移動させ，
  運転シミュレーションなどを体感できるものとなっている\cite{7759468}．
  このように，エンドエフェクタがフレームにワイヤで接続され，ワイヤ駆動により空間を移動するものは数多く見られ，
  同様の構成を持つものとして，最大43Gの加速度が出せる超高速ワイヤ駆動ロボットFALCON\cite{525288}や，
  産業用に開発されたCDPRであるIPAnema\cite{6695742}，
  スポーツのスタジアムに導入され，カメラを競技フィールド上で空間移動させるSkyCam\cite{cone1985skycam}などがある．
  いずれも，複数本のワイヤによる力伝達を用いてエンドエフェクタの姿勢を制御している．
  しかし，ワイヤに伝達させる力を発生させるアクチュエータはエンドエフェクタを囲む外枠のフレームに固定されており，
  ロボットの行動範囲はその内側に必ず制限される．

  \begin{figure}[t]
    \begin{center}
      \includegraphics[width=1\columnwidth]{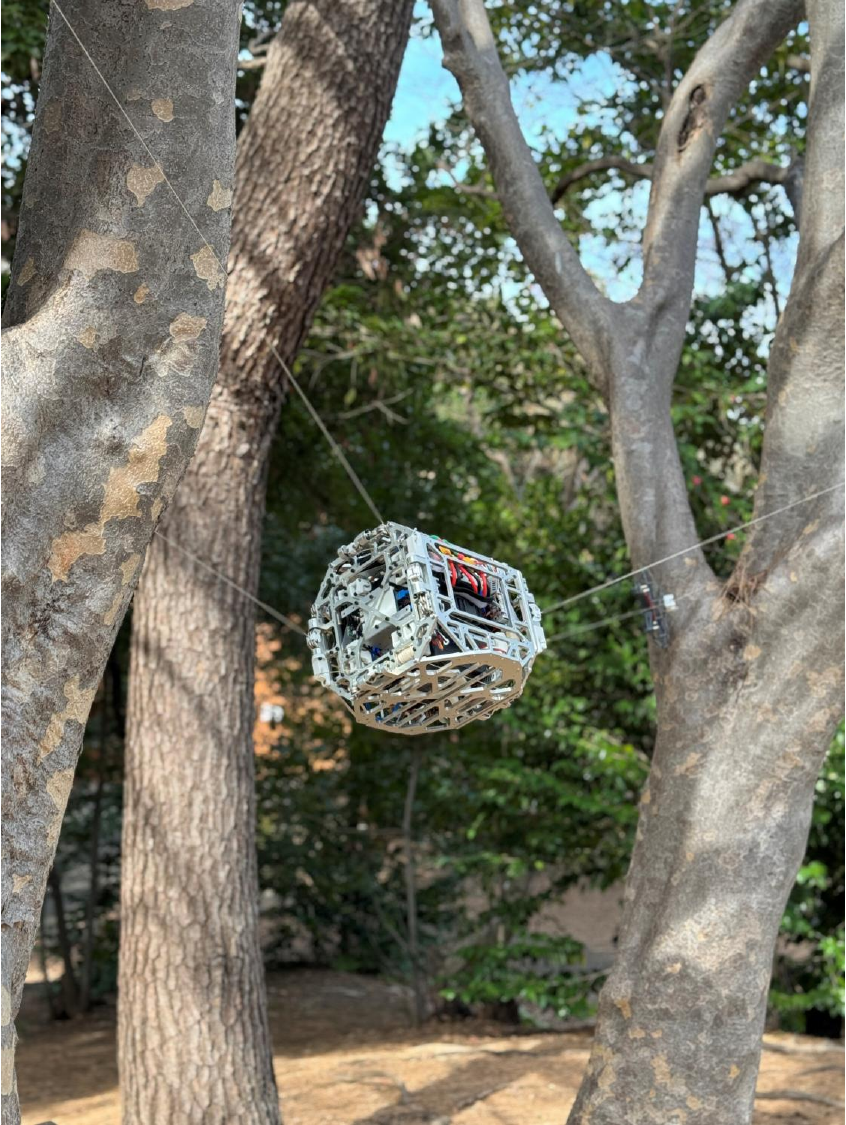}
      \vspace{-3.0ex}
      \caption{The overview of CubiX connecting to and utilizing the environment.}
      \vspace{-3.0ex}
      \label{fig:cubix}
    \end{center}
  \end{figure}
  % marionetの欠点をドローンで潰していたが，どうするの？ 
  % -> ある意味でポータブルだが，巨大なクレーンが必要でまったくポータブルでない
  % これに対して，ワイヤを巻き取るモジュールを搭載した台車を複数台用いることで，
  これに対して，ワイヤを巻き取るモジュールを搭載した複数台の台車をCDPRのフレームとすることで，
  CDPRがフレームごと移動することを可能にした研究\cite{xu2023collaborative,xu2023optimal,xu2024online}
  が存在するが，結果としてロボットの行動範囲はワイヤモジュールを運ぶ台車が移動可能な平面内に留まっている．
  また，ワイヤモジュールが取り付けられた巨大なクレーンを並べてCDPRを形成するMARIONET-CRANE\cite{5509299, merlet2010marionet}は，
  使用するワイヤの数だけ巨大なクレーンを人間が設置する必要があるため，
  % 人によるクレーンの設置を要する上に取り回しが悪く，
  CDPRがそのフレームにとらわれずに活動できるようになったとは言い難い．

  ここで，ワイヤを環境に接続して利用するロボットに注目すると，
  % さらに，ワイヤを環境に接続して利用する研究として，
  ワイヤで接続されたドローンと車両が連携することで車両の走破性を向上させる研究\cite{8794265}や，
  環境に脱着可能なカラビナでワイヤを接続して巻き取り，
  ヒューマノイドが発揮できる力を強化する研究\cite{10375200}がある．
  いずれの研究においても，ワイヤ駆動は直線的な力の援助という用途に留まっているが，
  ワイヤを扱うロボットにとって，ワイヤを介して環境を利用することは有効な手段の1つであると考えられる．

  そこで，本研究では\figref{fig:cubix}に示される，ワイヤを巻き取るアクチュエータをロボット体内に内蔵し，
  ロボット体内から出る複数のワイヤを環境に接続させ，それを巻き取ることで駆動するCubiXを提案する．
  これにより，
  ロボットが活動場所で環境を利用したパラレルワイヤ駆動を形成することを実現する．
}%

\section{Design of CubiX} \label{sec:design}
\subsection{Design of the Overall Structure}
\switchlanguage%
{%
  The overall structure of CubiX and its hardware configuration are shown in \figref{fig:overall_structure}.
  \begin{figure}[tbp]
    \begin{center}
      \includegraphics[width=1\columnwidth]{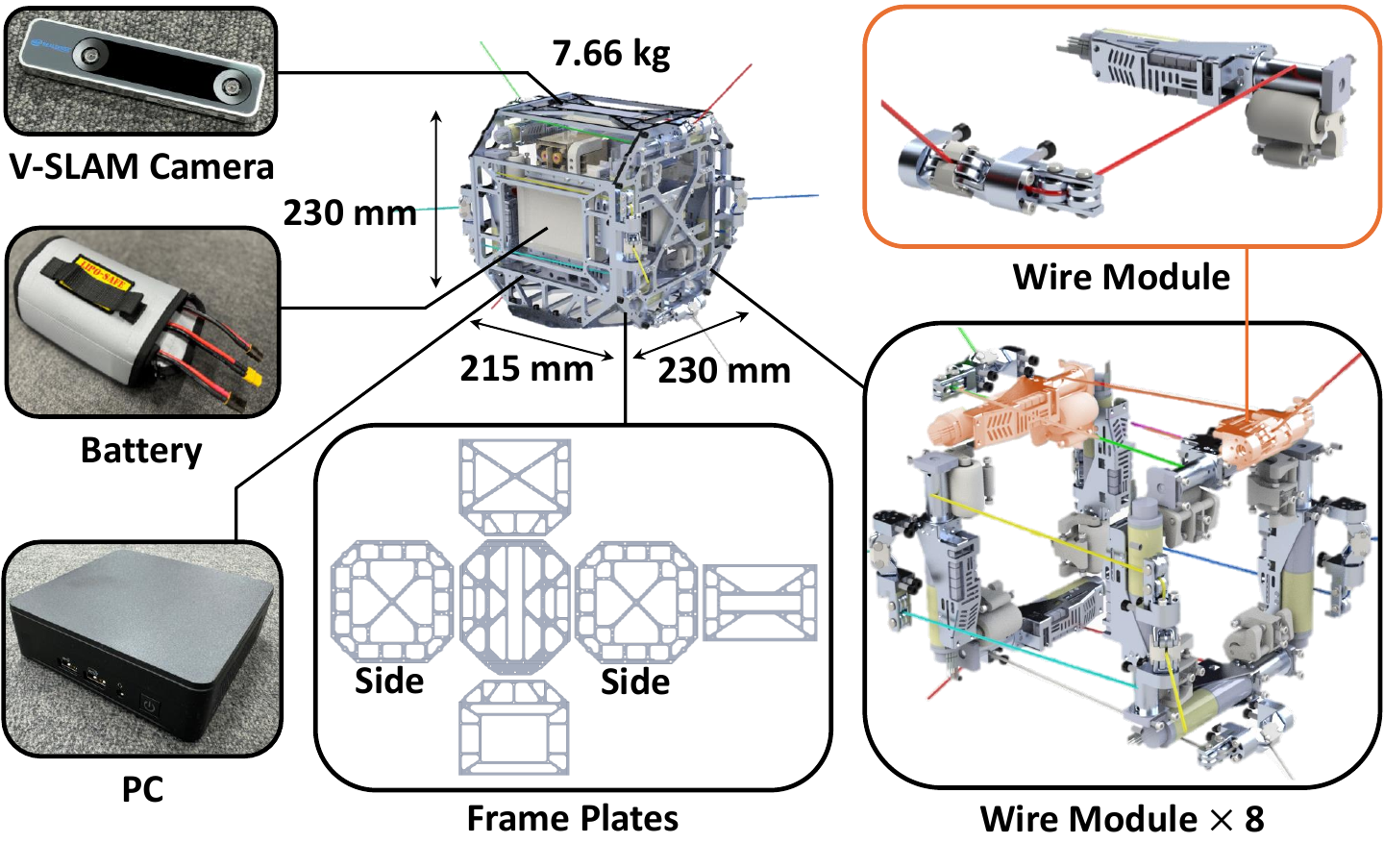}
      \vspace{-5.5ex}
      \caption{The hardware structure of CubiX. 
      It has 8 wire modules, and its cube-shaped structure consists of 6 frame plates.
      It is also equipped with devices such as a PC, battery, and camera necessary for operation.}
      \vspace{-5.5ex}
      \label{fig:overall_structure}
    \end{center}
  \end{figure}

  Wire-driven force can only be generated by winding wires, 
  and due to the property that unwinding wires does not produce force, the dimension controllable by $m$ wires is $m-1$ dimensional. 
  To control a 6-dimensional wrench combining 3-dimensional translational forces and 3-dimensional torques, 
  at least 7 wires are needed to generate driving force. 
  Considering CubiX's movement in 3-dimensional space, 
  it is desirable for CubiX to exert force in all directions from its central body, 
  thus requiring wires to be extended from CubiX in all directions. 
  Therefore, by also considering symmetry, a total of 8 wire modules are installed.

  As shown in the lower right of \figref{fig:overall_structure}, 
  the 8 wire modules are arranged along the edges of the cube.
  The wire path starts from the winches for winding the wire, 
  passes through wire relay points located on opposite edges, and exits CubiX's body. 
  This arrangement allows for a longer distance between the winch and the wire relay point, 
  reducing the fleet angle \cite{CHAPLIN199545}. 
  Moreover, since the wires only pass on the faces of the cube, batteries, circuit components, 
  and a PC can be installed inside CubiX without interference from wires.

  As shown in the lower center of \figref{fig:overall_structure}, 
  CubiX's cubic structure is composed of 6 frame plates. 
  4 wire modules are mounted on each of the 2 side frame plates, 
  and these 2 side frame plates are connected by the 4 middle frame plates. 
  Each frame plate has grooves, allowing the plates to interlock when assembling the cubic structure, thereby increasing its strength.
}%
{%
  CubiXの全体像と，ハードウェア構成を
  % それを構成するワイヤモジュール，フレームプレート，その他デバイスを
  \figref{fig:overall_structure}に示す．
  % CubiXには，8つのワイヤモジュールと，稼働に必要なセンサ，バッテリ，PCなどのデバイスを体内に搭載しており，
  % キューブ構造は6つのフレームプレートで構成される．
  \begin{figure}[tbp]
    \begin{center}
      \includegraphics[width=1\columnwidth]{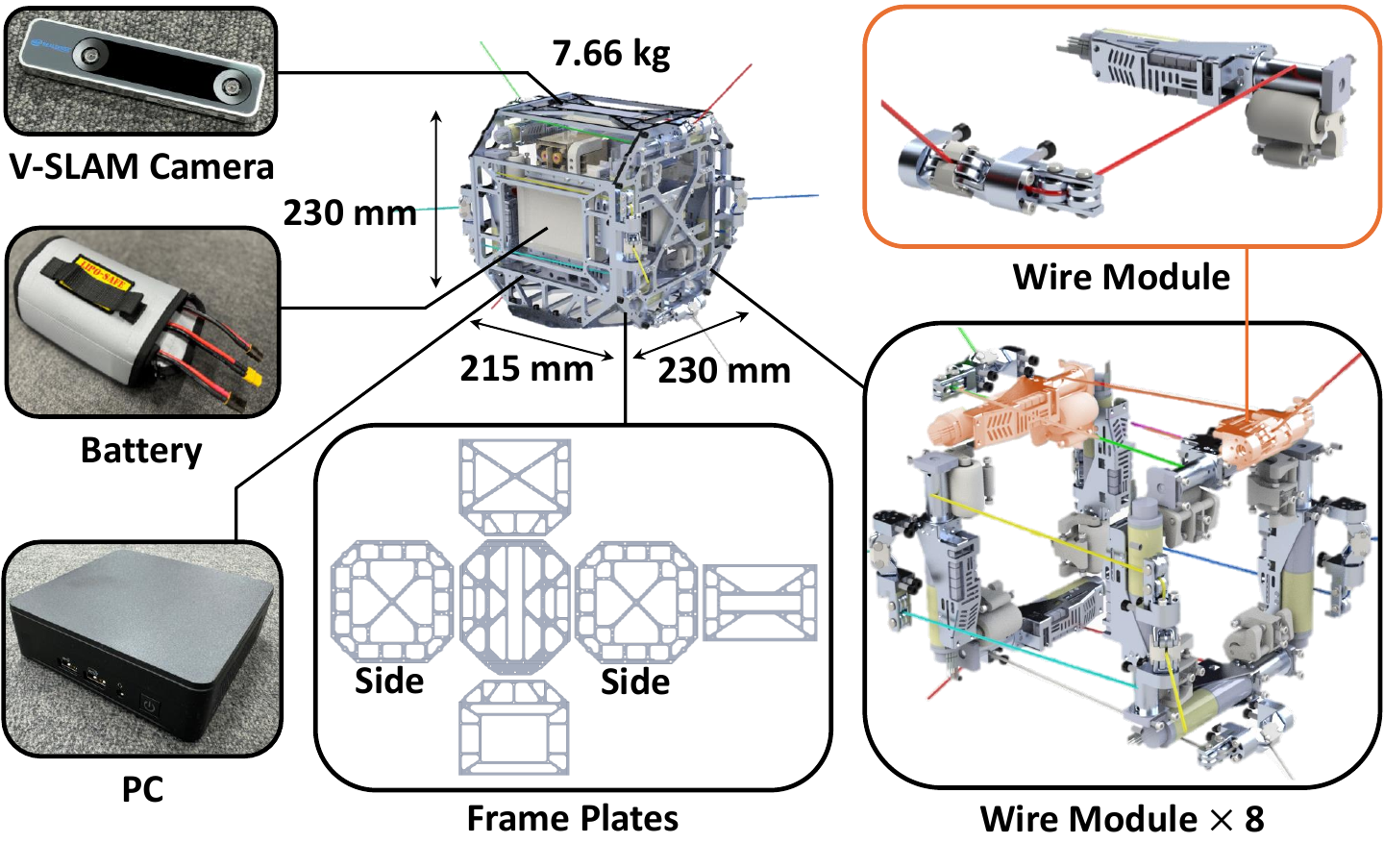}
      \vspace{-3.0ex}
      \caption{The hardware structure of CubiX. 
      It has 8 wire modules, and its cube-shaped structure consists of 6 frame plates.
      It is also equipped with devices such as a PC, battery, and camera necessary for operation.}
      \vspace{-3.0ex}
      \label{fig:overall_structure}
    \end{center}
  \end{figure}

  ワイヤ駆動はワイヤを巻き取ることでのみ駆動力を発生することができ，
  ワイヤを巻き出すことでは駆動力を発揮できないという性質上，$m$本のワイヤで制御できる次元は$m-1$次元である．
  つまり，3次元の並進力と3次元のトルクとを合わせた6次元のレンチを制御するためには，
  最低7本のワイヤにより駆動力を得る必要があり，CubiXにはそれ以上のワイヤモジュールを搭載する必要がある．
  また，CubiXが3次元空間を移動することを考えると，CubiX本体を中心としたあらゆる方向へ力が出せることが望ましいため，
  CubiXからワイヤが全方位に引き出せることが求められる．
  そこで，対称性を考えることで合計8つのワイヤモジュールを搭載した．

  \figref{fig:overall_structure}中の右下に示されるように，
  8つのワイヤモジュールはキューブ側面の辺に沿って配置されている．
  ワイヤの経路は，ワイヤを巻き取るプーリを出発し，それと向かい合う辺に配置されたワイヤ経由点を通って
  CubiXの体外へと出るというものである．
  このような配置をとることによって，ワイヤを巻き取るプーリとワイヤ経由点との距離を長くとることができ，
  フリートアングル\cite{CHAPLIN199545}を小さくすることにつながる．
  また，キューブの面上にのみワイヤが通ることになり，CubiX体内にバッテリや回路部品，制御PCをワイヤと干渉することなく搭載できる．

  \figref{fig:overall_structure}中の中央下に示されるように，
  CubiXのキューブ型構造は6面のフレームプレートによって成り立っている．
  側面の2面のフレームプレートにそれぞれ4つずつワイヤモジュールを組付けており，
  その2面を間の4面のフレームプレートによって接続している．
  各フレームプレートには溝が掘られており，キューブ型構造へ組み立てる際に
  各フレームプレート同士が溝にはめあうようにすることで，キューブ型構造の強度を上昇させている．
}%

\subsection{Design of the Wire Module}
\switchlanguage%
{%
  The overall view and individual components of the wire module of CubiX are shown in \figref{fig:wire_module}. 
  The wire module is comprised of three elements: the wire winding winch, the wire restraining pulley, and the wire relay unit, 
  designed to wind the wires connected to the environment and generate tension.

  The wire winding winch, shown in the upper left of \figref{fig:wire_module}, 
  winds the wire by rotating the winch connected to a motor. 
  Since CubiX incorporates a total of 8 wire modules, each wire winding winch is required to be lightweight and compact.
  Taking note of Musashi, a musculoskeletal humanoid equipped with 74 modules called muscle modules, 
  comprising a wire winding mechanism, motor, gear reducer, tension sensor, motor driver, and IMU \cite{kawaharazuka2019musashi}, 
  we referenced this design for the wire winding winch of CubiX.
  % Inspired by skeletal humanoid robot, Musashi \cite{kawaharazuka2019musashi}, 
  % which integrates over 100 modules known as muscle modules,
  % encompassing wire winding mechanisms, motors, gear reducers, tension sensors, motor drivers, and IMUs, 
  % CubiX's wire winding winches were designed. 
  These winches feature a torque constant of 14 mNm/A, a reduction ratio of 53:1 with a planetary gear head, 
  and dimensions of 16 mm in diameter and 35 mm in length, 
  achieving the performance parameters outlined in \tabref{tb:pulley_params}. 
  \begin{figure}[tbp]
    \begin{center}
      \includegraphics[width=1\columnwidth]{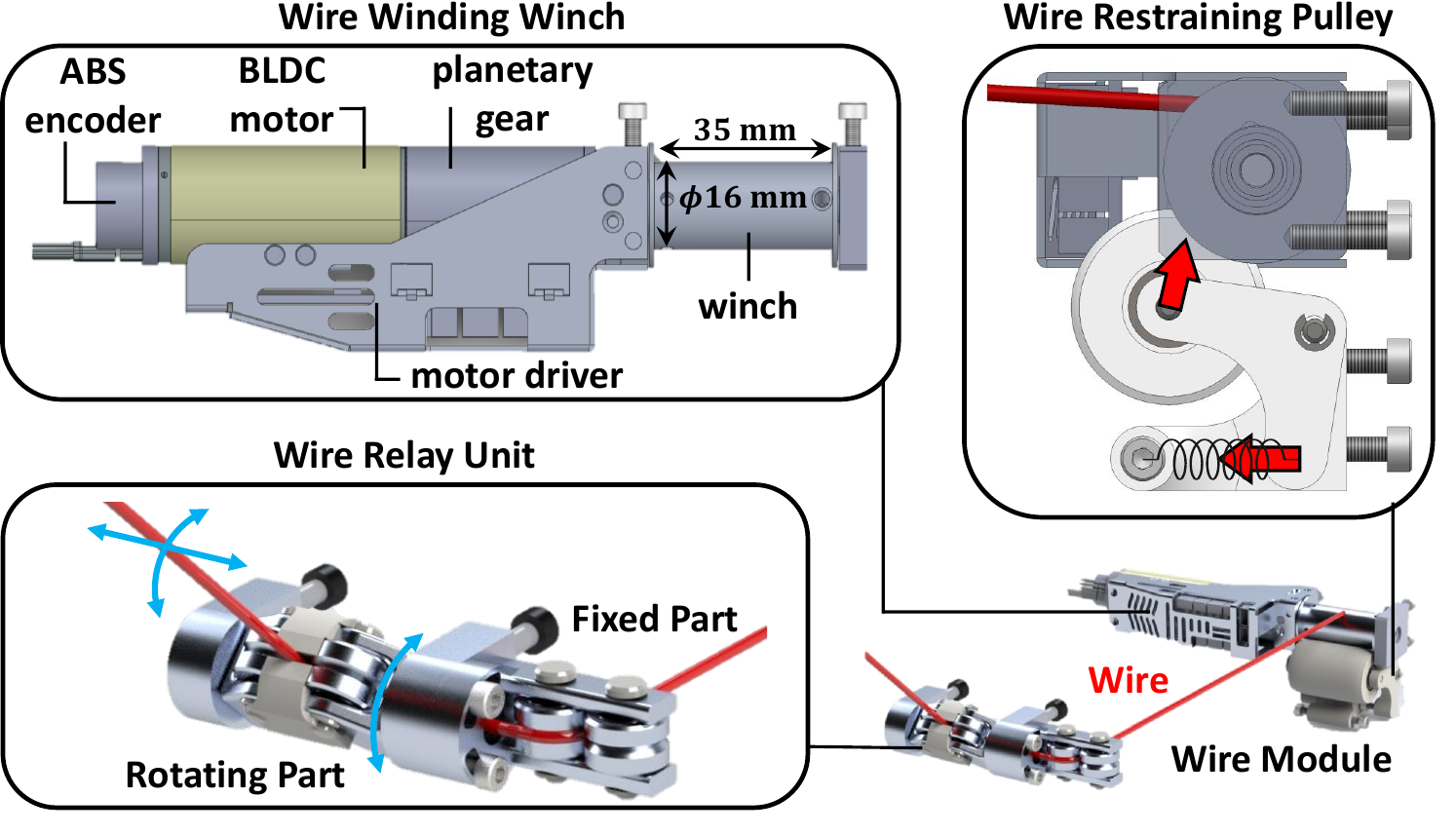}
      \vspace{-5.0ex}
      \caption{The overall structure of the wire module.
      It consists of 3 elements, the wire winding winch, the wire restraining pulley, and the wire relay unit.} 
      \vspace{-6ex}
      \label{fig:wire_module}
    \end{center}
  \end{figure}
  \begin{table}[htbp]
    % \centering
    \begin{center}
    \vspace{-2.0ex}
    \caption{Performance of the wire winding winch}
    \begin{tabular}{cc} \toprule
       Parameter & Value \\ \midrule
       Maximum Continuous Tension & 180 N \\ 
       Wire Winding Speed & 242 mm/s \\
       Wire Winding Length & 5.3 m \\ \bottomrule
    \end{tabular}
    \label{tb:pulley_params}
    \end{center}
    \vspace{-4.0ex}
  \end{table}
  \\The wires used are Vectran\textregistered, 
  high-performance ropes with a diameter of approximately 1.0 mm and a breaking strength of 1000 N.

  The wire restraining pulley, shown in the upper right of \figref{fig:wire_module}, 
  prevents irregular wire winding while the wire is being wound. 
  To wind up a long wire, multiple layers of wire are wound around the wire winding winch.
  The wire restraining pulley, pressed against the wire winding winch by tension springs, 
  prevents the wire from changing direction before the wire reaches the edge of the winding winch.
  Additionally, by smoothing the end face of the wire restraining pulley to the diameter of the wire, 
  the direction of winding is encouraged to switch when the wire reaches the end of the winding winch.
  Moreover, it prevents the wire from detaching from the wire winding winch when not under tension.

  The wire relay unit, shown in the lower left of \figref{fig:wire_module},
  aligns wires connected to the environment and transfers them to the wire winding winch. 
  Inspired by the wire-interference-driven robotic arm SAQIEL \cite{temma2024saqiel}, 
  which developed a mechanism for transferring wires between arbitrary points, 
  we used a similar mechanism in the wire relay unit of CubiX.
  Wires connected to the environment are transferred through the wire relay unit to the wire winding winch.
  At this time, 
  the rotational part of the wire relay unit rotates according to the direction of the wires connected to the environment.
  Thus, regardless of the orientation of the wire connection, the wire is aligned and transferred to the wire winding winch. 
  Consequently, regardless of CubiX's orientation, 
  wires connected to the environment are transferred in an aligned state to the wire winding winch.
  \begin{figure}[tbp]
    \begin{center}
      \includegraphics[width=1\columnwidth]{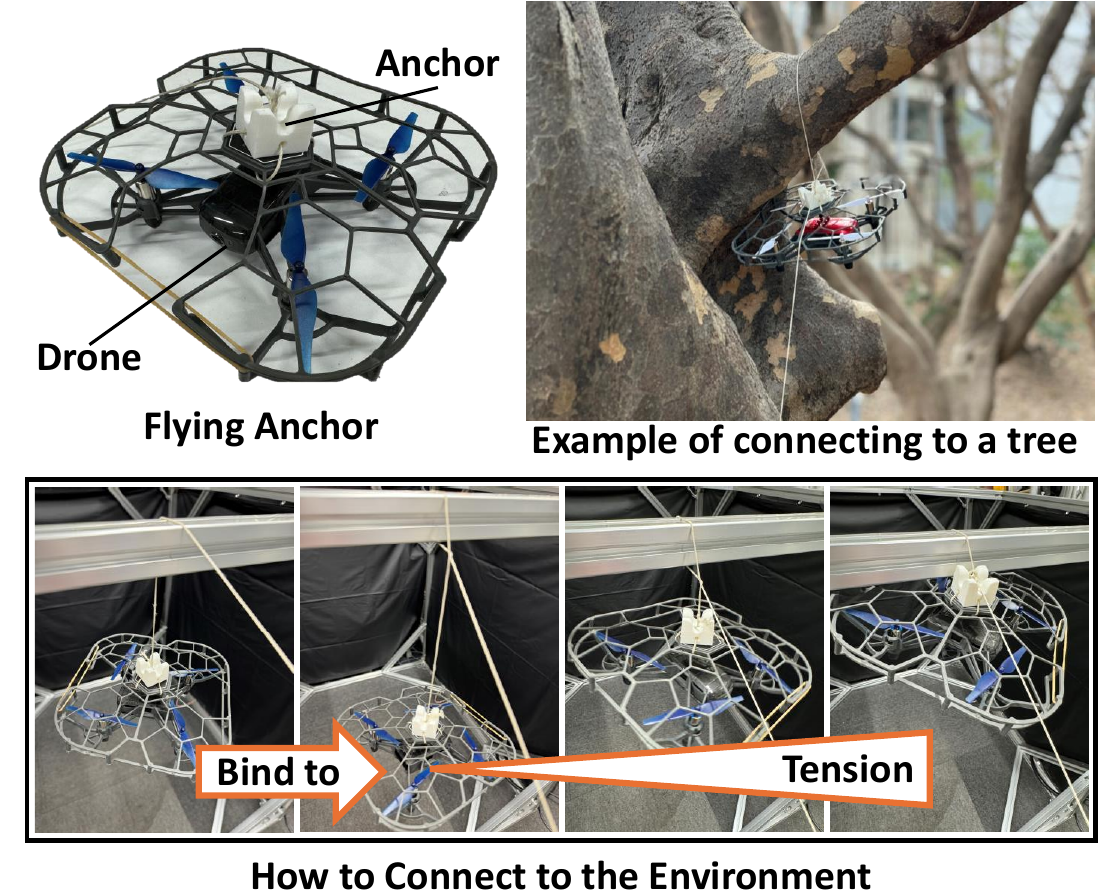}
      \vspace{-4.0ex}
      \caption{The overall structure of the flying anchor.
      The drone with the anchor circles around a pillar, looping the wire around it, 
      and then CubiX applies tension to anchor the wire to the environment.} 
      \vspace{-5.5ex}
      \label{fig:flying_anchor}
    \end{center}
  \end{figure}
}%
{%
  ワイヤモジュールの全体図と各機能を\figref{fig:wire_module}に示す．
  ワイヤモジュールは巻取りプーリ，押さえプーリ，ワイヤ経由点ユニットの3つの要素によって構成され，
  環境に接続されたワイヤを巻き取って張力を発生させるモジュールである．

  \figref{fig:wire_module}中の左上に示した巻取りプーリは，モータに接続したプーリを回転させることによりワイヤを巻き取る．
  CubiXには計8個のワイヤモジュールを搭載するため，1つ1つの巻取りプーリは軽量でコンパクトであることが求められる．
  筋骨格ヒューマノイドであるMusashi\cite{kawaharazuka2019musashi}には筋モジュールと呼ばれる，
  ワイヤ巻取り機構，モータ，減速機，張力センサ，モータドライバやIMUをまとめたモジュールを
  全身に計100以上搭載することができていることに注目し，CubiXの巻取りプーリの構成においてもこれを参考にした．
  トルク定数14 mNm/AのBLDCモータに，減速比53:1のプラネタリギアヘッド，
  直径16 mm，長さ35 mmのプーリを取り付けてワイヤを巻き取ることで，\tabref{tb:pulley_params}に示す
  性能を持つ巻取りプーリを実現した．
  \begin{figure}[tbp]
    \begin{center}
      \includegraphics[width=1\columnwidth]{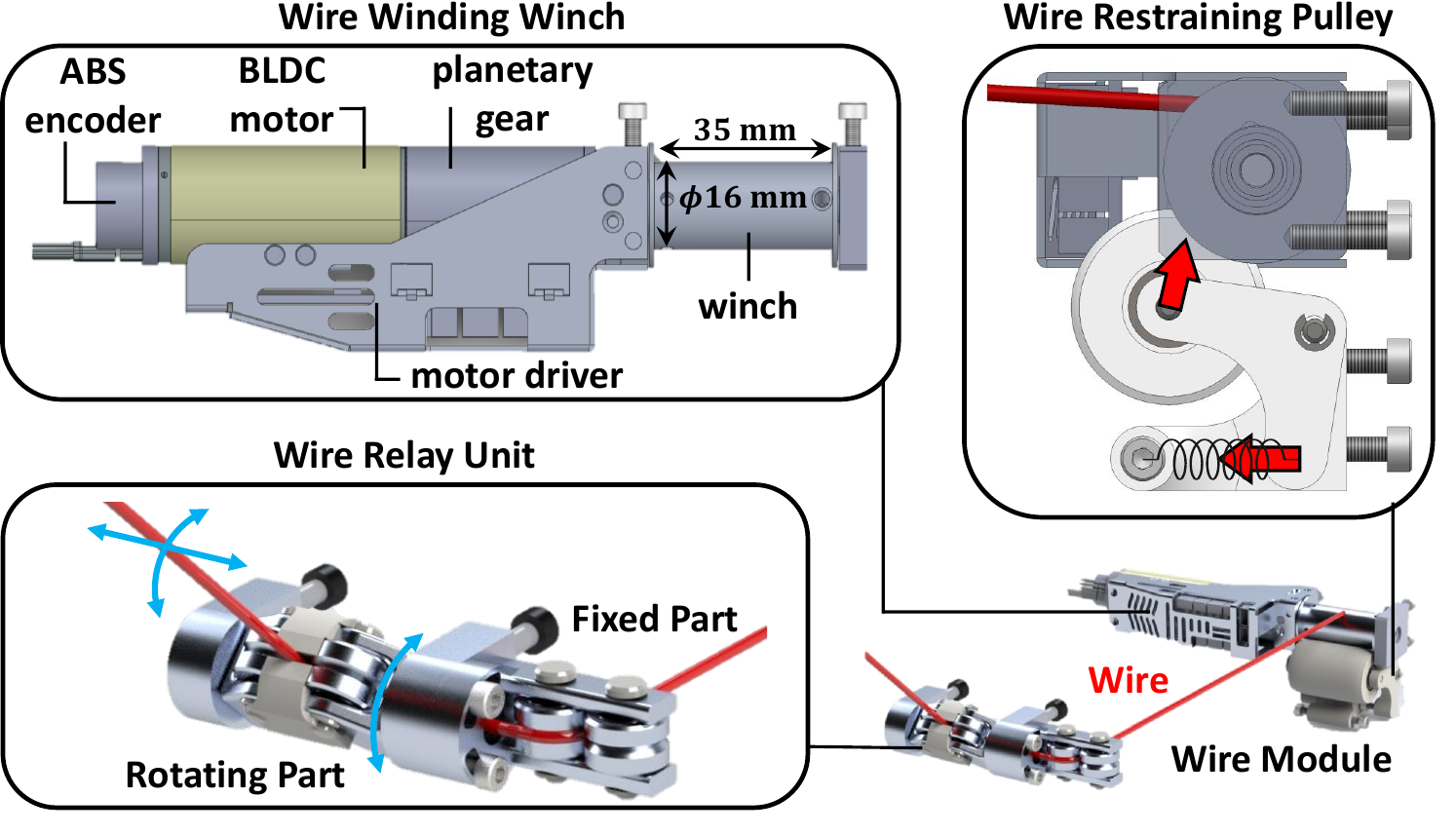}
      \vspace{-3.0ex}
      \caption{The overall structure of the wire module.
      It consists of 3 elements, Wire Winding Pulley, Wire Restraining Pulley, and Wire Relay Unit.} 
      \vspace{-3.0ex}
      \label{fig:wire_module}
    \end{center}
  \end{figure}
  \begin{table}[htbp]
    % \centering
    \begin{center}
    \begin{tabular}{cc} \toprule
       Parameter & Value \\ \midrule
       Maximum Continuous Tension & 180 N \\ 
       Wire Winding Speed & 242 mm/s \\
       Wire Winding Length & 5.3 m \\ \bottomrule
    \end{tabular}
    \caption{Performance of winding winch}
    \label{tb:pulley_params}
    \end{center}
    \vspace{-3.0ex}
  \end{table}
  \\なお，巻き取るワイヤにはVectran
  \textregistered
  という，高強力ポリアレート繊維を使用した高性能ロープで線径が約1.0 mm，破断強度が1000 Nのものを採用した．

  \figref{fig:wire_module}中の右上に示した押さえプーリは，
  巻取りプーリがワイヤを巻き取る際，それが乱巻きになることを抑制する．%役割を果たす．
  CubiXのワイヤモジュールでは，長いワイヤを巻き取るために，
  巻取りプーリに多重にワイヤを巻き取る．
  押さえプーリを引張りバネによって巻取りプーリに押し当てることで，
  % 巻取りプーリの端までワイヤが巻き取られることを促す．
  ワイヤが巻き取られていく方向が巻取りプーリの端に到達する前に切り替わることを抑制する．
  また，押さえプーリの端面をワイヤ直径分なめらかにすぼませることで，
  巻取りプーリの端にワイヤが到達したときに巻き取り方向が切り替わることを促す．
  % 引張りバネは初張力1.57N，ばね定数1.27N/mmであり，組み付けられた状態で9mm伸ばされているため，
  % 押さえプーリの両端をそれぞれ13Nで巻取りプーリへ押し当てている．
  また，ワイヤに張力がかけられていないときにワイヤが巻取りプーリから解け離れることも抑制する．%できている．

  \figref{fig:wire_module}中の左下に示したワイヤ経由点ユニットは，
  環境へ接続されたワイヤを整列し，巻取りプーリへ受け渡す役割を持つ．
  ワイヤ干渉駆動ロボットアームであるSAQIEL\cite{temma2024saqiel}にて，
  ワイヤを任意の二点間で受け渡す機構が開発され，
  CubiXのワイヤ経由点ユニットはその機構と同様のものを使用した．
  環境に接続されたワイヤはワイヤ経由点ユニットを介して%の回転部から固定部に受け渡され， その後に
  巻取りプーリへ渡される．
  このとき，ワイヤ経由点ユニットの回転部は，環境に接続されたワイヤの方向に従って回転する．
  よって，どのような向きにワイヤが接続されていても，ワイヤは整列され巻取りプーリへ受け渡される．
  % この際，回転部は環境に接続されたワイヤの方向にしたがって回転することで，
  % CubiXからみた環境に接続されたワイヤ端点の相対位置がどのような場合でも，
  % ワイヤは整列され巻取りプーリへ受け渡される．
  つまり，CubiXがどのような姿勢であっても，環境に接続されたワイヤは整列した状態で巻取りプーリに受け渡される．
}%

\subsection{Electrical Configuration}
\switchlanguage%
{%
  The devices installed in CubiX are listed in \tabref{tb:devices}.
  \begin{table}[htbp]
    \centering
    \vspace{-1.0ex}
    \caption{Devices installed into CubiX}
    \vspace{-2.0ex}
    \begin{tabular}{|l||c|c|c|}
      \hline
      Device & Description & Quantity\\ \hline\hline
      PC & Intel NUC 12 Pro Mini PC Kit& 1 \\ \hline
      % Ethernet Relay Board & ENB-05\cite{D-thesis:Nagamatsu:2021} & 1 \\ \hline
      % Motor Driver & FPGA-03D\cite{D-thesis:Nagamatsu:2021} & 8 \\ \hline
      Wireless Emergency & Harmony ZBRRA & 1 \\ 
      Stop Receiver &  &  \\ \hline
      Power Relay & OMRON G9EA-1-B DC24 & 1 \\ \hline
      Camera & Intel RealSense & \\
       & Tracking Camera T265 & 1 \\ \hline
      Logic Battery & HRB 3S 6000mAh 11.1V & 1 \\ \hline
      Power Battery & HRB 6S 3300mAh 22.2V & 2 \\ \hline
    \end{tabular}
    \vspace{-2.0ex}
    \label{tb:devices}
  \end{table}
  \\CubiX is required to operate autonomously; therefore, devices necessary for operation such as PC, battery, 
  and sensors are installed inside its body.
  The interior of CubiX is divided into three tiers using sheet metal parts: the lower tier houses the PC, 
  the middle tier contains the battery, and the upper tier accommodates other circuit devices including the V-SLAM camera.
  The battery voltage is 12V for the logic system and 48V for the power system.
}%
{%
  CubiXに搭載したデバイスは\tabref{tb:devices}の通りである．
  \begin{table}[htbp]
    \centering
    \begin{tabular}{|l||c|c|c|}
      \hline
      Device & Description & Quantity\\ \hline\hline
      PC & Intel NUC 12 Pro Mini PC Kit& 1 \\ \hline
      % Ethernet Relay Board & ENB-05\cite{D-thesis:Nagamatsu:2021} & 1 \\ \hline
      % Motor Driver & FPGA-03D\cite{D-thesis:Nagamatsu:2021} & 8 \\ \hline
      Wireless Emergency & Harmony ZBRRA & 1 \\ 
      Stop Receiver &  &  \\ \hline
      Relay & OMRON G9EA-1-B DC24 & 1 \\ \hline
      Camera & Intel RealSense & \\
       & Tracking Camera T265 & 1 \\ \hline
      Logic Battery & HRB 3S 6000mAh 11.1V & 1 \\ \hline
      Power Battery & HRB 6S 3300mAh 22.2V & 2 \\ \hline
    \end{tabular}
    \caption{Devices integrated into CubiX}
    \vspace{-3.0ex}
    \label{tb:devices}
  \end{table}
  \\CubiXは自律して活動することが求められるため，
  PC，バッテリ，センサといった稼働に必要なデバイスをその体内に搭載した．
  CubiXの体内を板金パーツによって3段に分割し，下段にはPC，中段にはバッテリ，上段にはその他の回路デバイスを配置した．
  バッテリ電圧はロジック系統が12V，パワー系統が48Vである．
  % モータドライバにはモータエンコーダを用いた分解能の細かいベクトル制御や
  % 基板単体で1ms周期のサーボ制御が実装されているFPGA-03D\cite{D-thesis:Nagamatsu:2021}を搭載した．
  % また，モータドライバとPCとのEthernet通信を中継する基板であるENB-05\cite{D-thesis:Nagamatsu:2021}により，
  % データ転送レート1Gbpsの通信をモータドライバとPCとの間で実現している．
}%

\subsection{Flying Anchor to Connect Wires to Environment}
\switchlanguage%
{%
  There are various means to connect wires to the environment, 
  such as manual tying by human hands, using carabiners for connection, or throwing hooks to attach to objects.
  As an example of these methods, we present a flying anchor using a drone, as shown in \figref{fig:flying_anchor}.
  The drone winds the wire around objects like pillars or trees in the environment and secures the wound wire to the environment 
  using the anchor.

  For instance, when connecting the wire to a pillar, as shown in the 4 images in \figref{fig:flying_anchor}, 
  the drone at the end of the wire circles around the pillar to tie the wire. By applying tension to the wire in this state, 
  the anchor attaches to the wire tied to the pillar, securing the wire to the environment. 
  To undo this, loosening the tension releases the anchor from the knot using the drone's weight. 
  At this point, with the anchor attached to the top center of the drone, the drone's posture becomes horizontal in mid-air, 
  allowing it to fly again to untie the wire. 
  The anchor, despite its lightweight at 5 g, has been confirmed to withstand a load of up to 35 kg. 
  Additionally, a lightweight and compact drone, Tello EDU, was used for this purpose.
}%
{%
  \begin{figure}[tbp]
    \begin{center}
      \includegraphics[width=1\columnwidth]{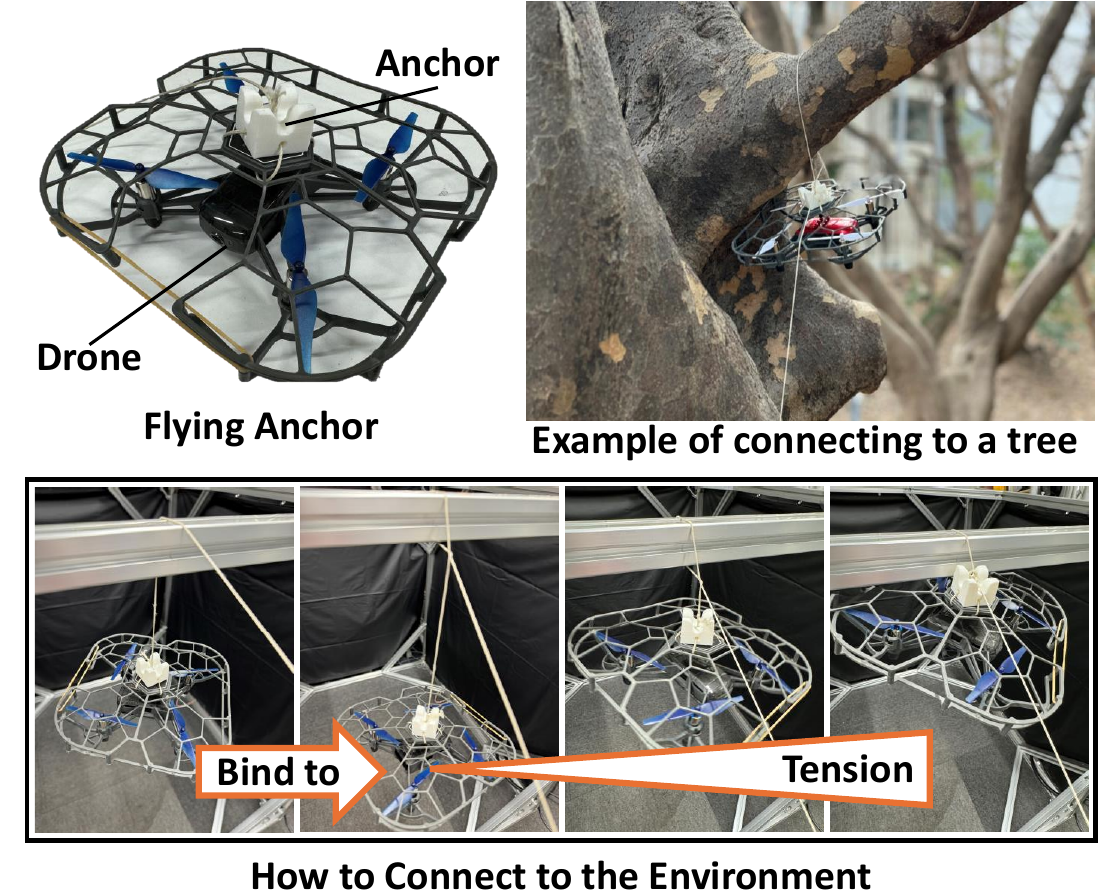}
      \vspace{-3.0ex}
      \caption{The overall structure of the flying anchor.
      The drone with the anchor circles around a pillar, looping the wire around it, 
      and then CubiX applies tension to anchor the wire to the environment.} 
      \vspace{-3.0ex}
      \label{fig:flying_anchor}
    \end{center}
  \end{figure}
  ワイヤを環境に接続する手段は，
  人の手による締結や，カラビナを用いた接続，
  フックを投擲して引っ掛けるなど
  様々あるが，本研究ではその手段の一例として，
  % ワイヤを環境に接続する手段の一例として，
  ドローンを用いた飛行アンカーを\figref{fig:flying_anchor}に示す．
  ドローンによって柱や木などの環境へワイヤを巻き付け，アンカーによって巻き付けたワイヤを環境へ縛り付ける．

  例えば，柱にワイヤを縛り付ける際には，\figref{fig:flying_anchor}中の4枚の画像で示されているように，
  % アンカーを機体上面に搭載したドローンが，
  ワイヤの先端にあるドローンが柱を一周し，ワイヤを結ぶ．
  その状態でワイヤに張力をかけることで，
  柱に結ばれたワイヤにアンカーが掛かり，ワイヤが環境に縛り付けられる．
  % 環境にワイヤを一周させて交差させ，
  % その状態でワイヤに張力をかけると，
  % 柱で交差するワイヤにアンカーが掛かり，ワイヤが環境に縛り付けられる．
  これを解く際には，かけた張力を緩めることで，ドローンの重力を用いてアンカーを結び目から離す．
  % かけた張力を緩めるとドローンの重量でアンカーが結び目から離れることで噛みつきを解除することができる．
  このとき，アンカーがドローンの上面中心に取り付けられていることで姿勢が空中で水平となり，
  そこから再び飛行することでワイヤを解く動作ができる．
  アンカーは5 gと軽量でありながら35 kgの荷重に耐えることが確認された．
  なお，ドローンには軽量かつコンパクトなTello EDUを使用した．
}%

\section{Controller of CubiX} \label{sec:controller}
\subsection{Pose Control of Wire-driven Floating Link}
\switchlanguage%
{%
  \begin{figure}[tbp]
    \begin{center}
      \includegraphics[width=1.0\columnwidth]{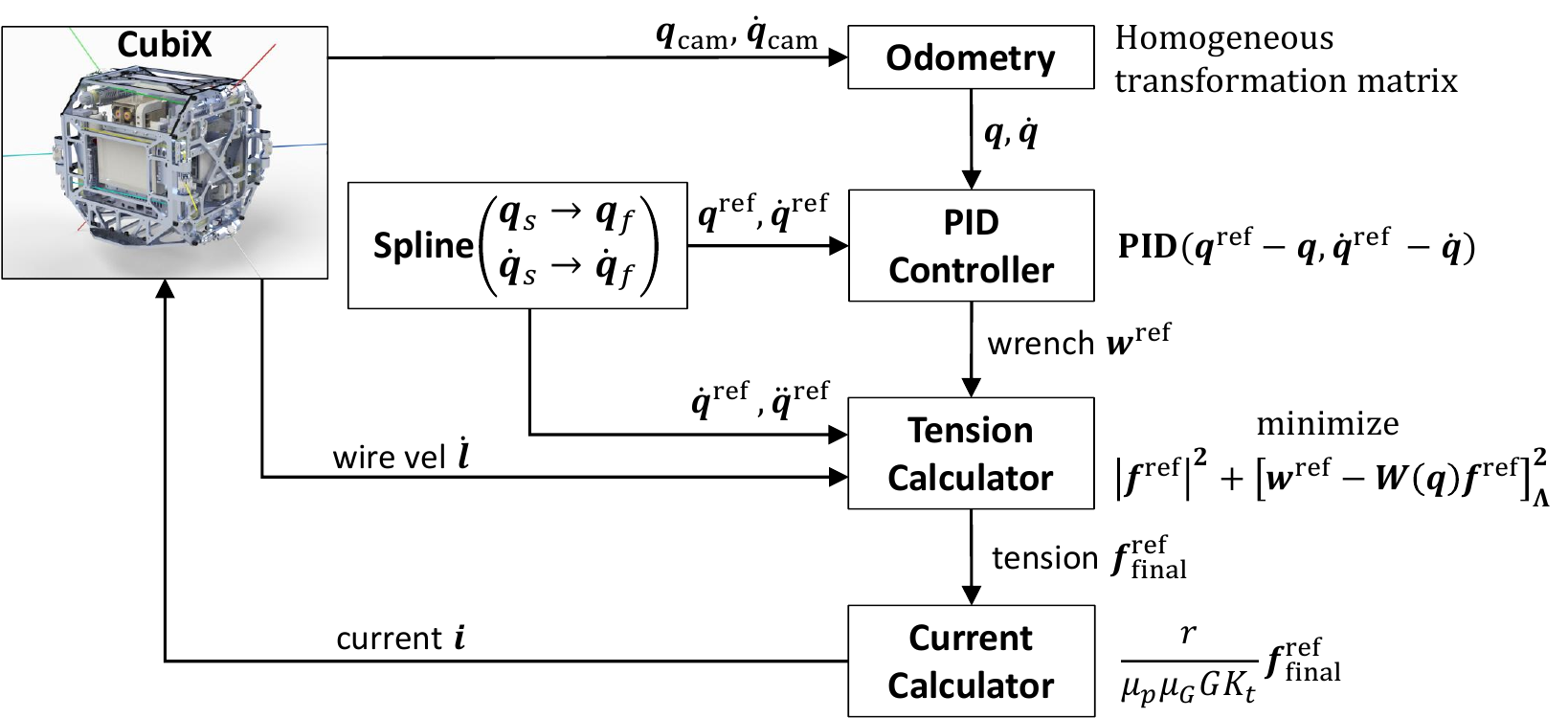} 
      \vspace{-3.0ex}
      \caption{Pose control loop to move the pose of CubiX to the target pose.
        Calculate the tension required for each wire to follow the spline path.} 
      \vspace{-5.0ex}
      \label{fig:position_control_flow}
    \end{center}
  \end{figure}
  CubiX is considered as a wire-driven floating link connected to the environment via a free joint, 
  allowing it to move in 3-dimensional space. 
  The pose control loop of CubiX is shown in \figref{fig:position_control_flow}.

  In the Odometry block of \figref{fig:position_control_flow}, 
  the pose $\bm{q}_{\mathrm{cam}}$ and velocity $\bm{\dot{q}}_{\mathrm{cam}}$
  estimated by the Intel RealSense Tracking Camera T265 using visual SLAM
  are transformed into the coordinates of CubiX's center,
  yielding $\bm{q}$ and $\bm{\dot{q}}$.

  In the Spline block of \figref{fig:position_control_flow}, 
  a 3rd-order spline path is computed to reach the specified final pose and velocity $\bm{q}_{f},\;\bm{\dot{q}}_{f}$ 
  from the initial pose and velocity $\bm{q}_{s},\;\bm{\dot{q}}_{s}$ within a designated time.
  This results in obtaining the target pose $\bm{q}^\mathrm{ref}$, target velocity $\bm{\dot{q}}^\mathrm{ref}$, 
  and target acceleration $\bm{\ddot{q}}^\mathrm{ref}$.

  In the PID Controller block of \figref{fig:position_control_flow}, 
  a PID control is performed to adjust $\bm{q},\;\bm{\dot{q}}$ to match $\bm{q}^\mathrm{ref},\;\bm{\dot{q}}^\mathrm{ref}$, 
  resulting in the feedback wrench $\bm{w}_{\mathrm{fb}}$. 
  Additionally, the gravity compensation wrench $\bm{w}_{g}$ 
  calculated by Pinocchio \cite{carpentier-sii19}, the dynamics computation library,
  is added as a feedforward wrench to obtain the desired wrench $\bm{w}^\mathrm{ref}$.

  \begin{figure}[tbp]
    \begin{center}
      \includegraphics[width=1.0\columnwidth]{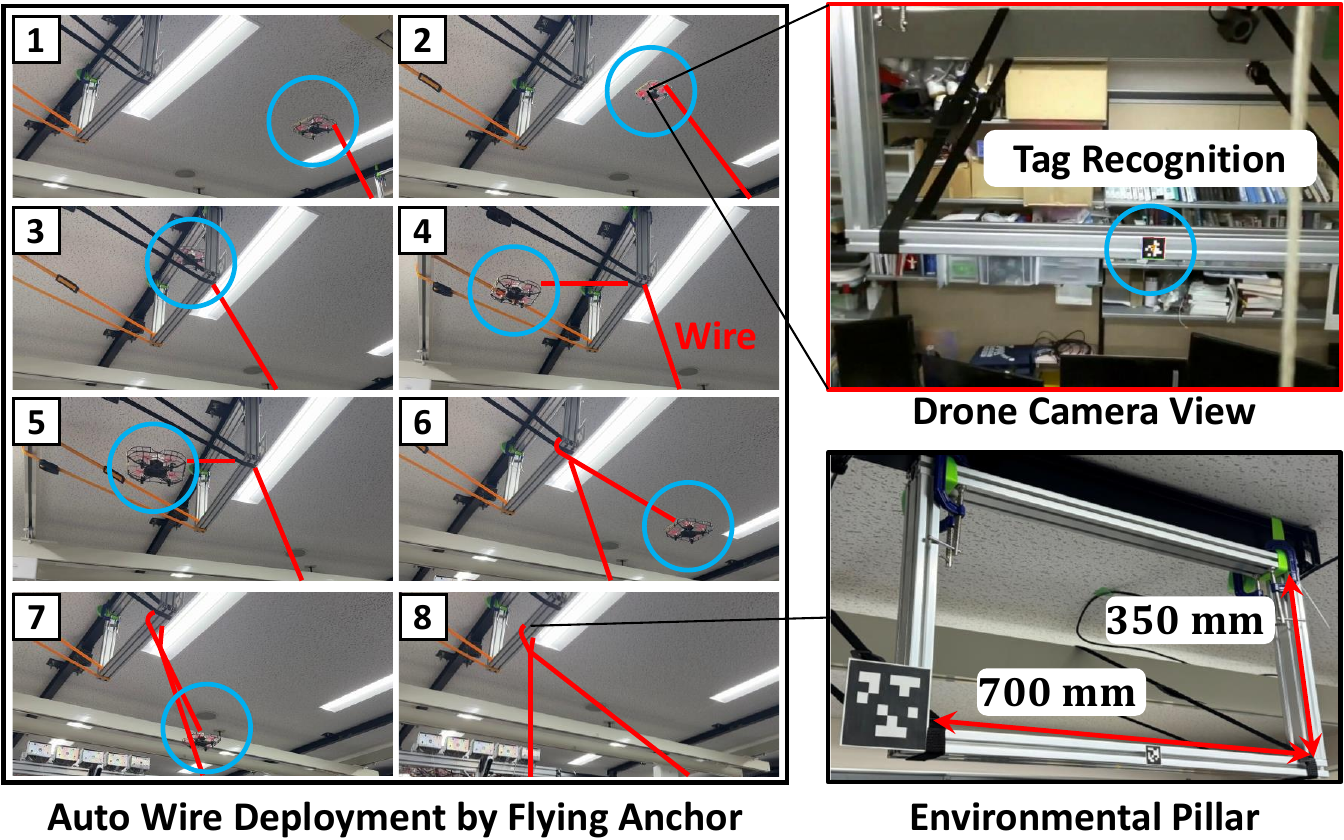} 
      \vspace{-4.0ex}
      \caption{System of wire deployment using the flying anchor.
      The relative pose of the flying anchor and the pillar is estimated using tag recognition, 
      and the wire is connected to the pillar following the given path.} 
      \vspace{-5.0ex}
      \label{fig:flying_anchor_placement}
    \end{center}
  \end{figure}

  Next, in the Tension Calculator block of \figref{fig:position_control_flow}, 
  $\bm{w}^\mathrm{ref}$ is converted into the tension $\bm{f}^\mathrm{ref}$ that each wire should exert. 
  The total number of wires is denoted as $m$, with $i\in{1, ..., m}$.
  The position where the $i$-th wire exits from CubiX, as viewed from the CubiX origin, is denoted by $\bm{r}_i$,
  and the unit vector pointing from its outlet toward the environment is denoted by $\bm{s}_i$.
  In this context, using the $6 \times m$ Jacobian matrix $\bm{W}(\bm{q})$, 
  the wrench $\bm{w}$ exerted by the tension $\bm{f}$ is expressed as shown in \equref{eq:wire_matrix}.
  Here, the tension $\bm{f}$ consists of the $i$-th wire tension $f_i$, where $\bm{f} = [f_1, ..., f_m]^\top$,
  and the tensile direction is considered positive.
  \begin{equation}
    \begin{split}
      \bm{w} &= {\bm{W}(\bm{q})}{\bm{f}},\;\; \mathrm{where}\;\;
      \bm{W}(\bm{q}) =
      \left[ \begin{array}{ccc}
        \bm{s}_{1} & \cdots & \bm{s}_{m} \\
        \bm{r}_{1}\times\bm{s}_{1} & \cdots & \bm{r}_{m}\times\bm{s}_{m}
      \end{array} \right]
    \end{split}
    \label{eq:wire_matrix}
  \end{equation}
  Based on this, a tension-based joint-space controller, which has been studied for musculoskeletal humanoid control \cite{7803367},
  is referenced to determine the tension $\bm{f}^\mathrm{ref}$ exerting the desired wrench $\bm{w}^\mathrm{ref}$
  by solving the quadratic programming problem in \equref{eq:quad_prog}.
  Here, $\bm{\Lambda}$ denotes the weight matrix, and $\bm{f}^\mathrm{min},\;\bm{f}^\mathrm{max}$ 
  represent the vectors of minimum and maximum wire tensions.
  \begin{equation}
    \begin{aligned}
      & \underset{\bm{f}^\mathrm{ref}}{\text{min}}
      & & \|\bm{f}^\mathrm{ref}\|^2 + \left[\bm{w}^\mathrm{ref} - {\bm{W}(\bm{q})}{\bm{f}^\mathrm{ref}}\right]^\top \bm{\Lambda} \left[\bm{w}^\mathrm{ref} - {\bm{W}(\bm{q})}{\bm{f}^\mathrm{ref}}\right]\\
      & \text{s.t.}
      & & \bm{f}^\mathrm{min} \leq \bm{f}^\mathrm{ref} \leq \bm{f}^\mathrm{max} \\
    \end{aligned}
    \label{eq:quad_prog}
  \end{equation}
  In addition, compensatory tensions are computed to offset wire winding winch inertia torque and shaft friction loss torque.
  These tensions are added to $\bm{f}^{\mathrm{ref}}$ to obtain $\bm{f}^\mathrm{ref}_{\mathrm{final}}$.
  These computations rely on acceleration $\bm{\ddot{q}}^{\mathrm{ref}}$ and wire velocity $\bm{\dot{l}}$.

  Finally, in the Current Calculator block of \figref{fig:position_control_flow}, 
  the current $\bm{i}$ commanded to each motor is obtained by scaling $\bm{f}^\mathrm{ref}_{\mathrm{final}}$ according to \equref{eq:current}, 
  where $r$ is the wire winding winch radius, $\mu_p$ is the pulley transmission efficiency, 
  $\mu_G$ is the gearhead transmission efficiency, $G$ is the reduction gear ratio, and $K_t$ is the torque constant.
  \begin{equation}
    \bm{i} = \cfrac{r}{\mu_p \mu_G G K_t}\bm{f}^\mathrm{ref}_\mathrm{final}
    \label{eq:current}
  \end{equation}

  Thus, wire tension-based pose control is achieved to enable CubiX to follow its path.
}%
{%
  CubiXは環境とワイヤによって接続され3次元空間上を移動するため，
  環境にフリージョイントで接続されたワイヤ駆動浮遊リンクであると考えられる．
  CubiXの位置制御の流れを\figref{fig:position_control_flow}に示す．
  \begin{figure}[tbp]
    \begin{center}
      \includegraphics[width=1.0\columnwidth]{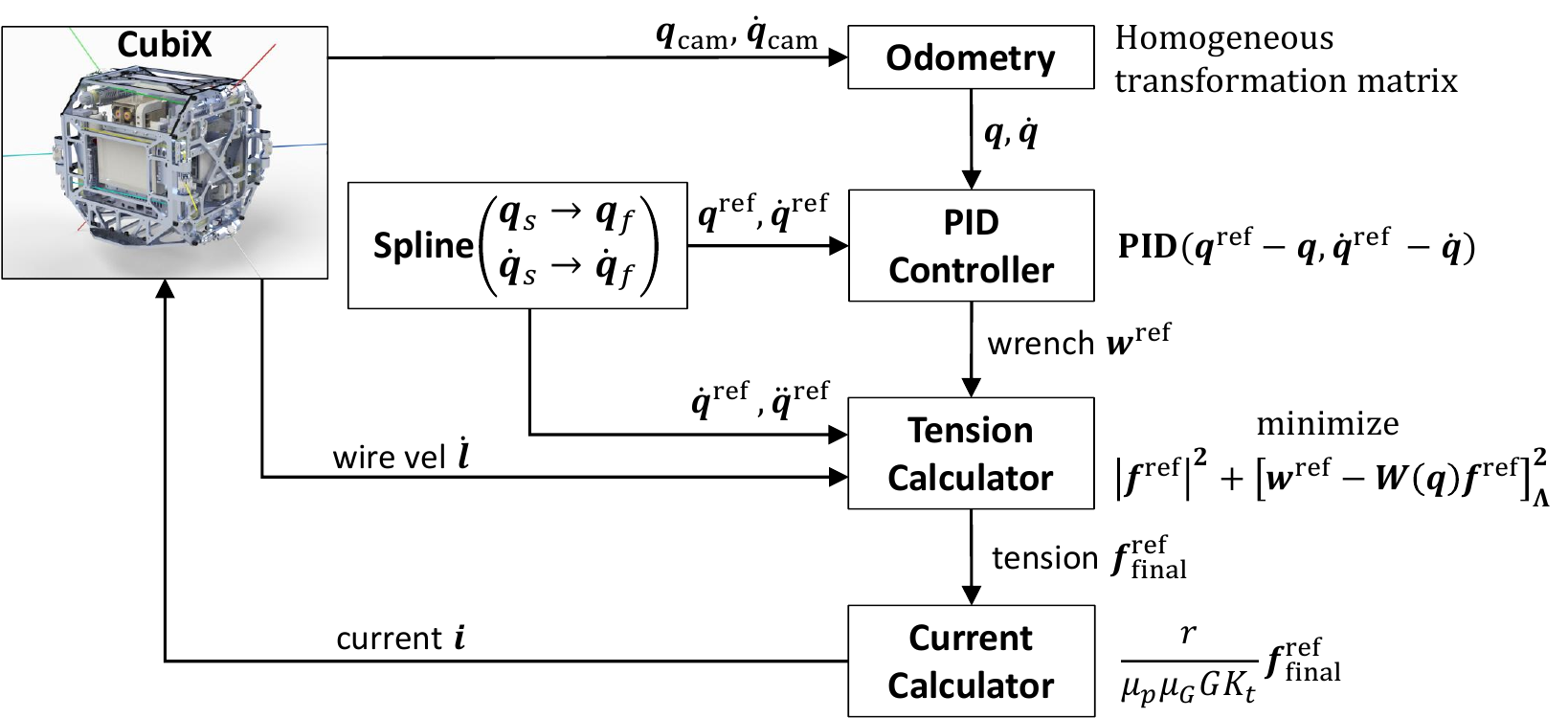} 
      \vspace{-3.0ex}
      \caption{Pose control flow to move the pose of CubiX to the target pose.
        Calculate the tension required for each wire to follow the spline path.} 
      \vspace{-3.0ex}
      \label{fig:position_control_flow}
    \end{center}
  \end{figure}

  \figref{fig:position_control_flow}中のOdometryでは，
  Intel社製のRealsense Tracking Camera T265
  % が内部に搭載したVPU
  によるビジュアルスラムにより推定された
  位置$\bm{q}_{\mathrm{cam}}$，
  速度$\bm{\dot{q}}_{\mathrm{cam}}$を座標変換することで，
  CubiXの中心座標における$\bm{q},\;\bm{\dot{q}}$が計算される．
  ここで，位置とは，並進位置と回転姿勢を合わせたベクトルである．

  \figref{fig:position_control_flow}中のSplineでは，
  初期位置$\bm{q}_{s}$，初期速度$\bm{\dot{q}}_{s}$から
  与えられた終了位置$\bm{q}_{f}$，終了速度$\bm{\dot{q}}_{f}$
  へ指定された時間で到達するような3次スプライン経路を計算する．%ことで，
  これにより，目標位置$\bm{q}^\mathrm{ref}$，目標速度$\bm{\dot{q}}^\mathrm{ref}$，目標加速度$\bm{\ddot{q}}^\mathrm{ref}$を得る．

  \figref{fig:position_control_flow}中のPID Controllerでは，
  $\bm{q},\;\bm{\dot{q}}$が$\bm{q}^\mathrm{ref},\;\bm{\dot{q}}^\mathrm{ref}$となるようにPID制御を行うことで
  フィーバックレンチ$\bm{w}_{\mathrm{fb}}$を計算する．
  これに，フィードフォワードレンチとして，
  動力学計算ライブラリであるpinocchio\cite{carpentier-sii19}により計算した
  重力補償レンチ$\bm{w}_{g}$を加えることで
  目標レンチ$\bm{w}^\mathrm{ref}$を得る．

  % パラレルワイヤ駆動ロボットの力学と制御の基礎\cite{kino2021wire}を参考に，
  次に，
  \figref{fig:position_control_flow}中のTension Calculatorで
  $\bm{w}^\mathrm{ref}$を各ワイヤで発揮するべき張力である$\bm{f}^\mathrm{ref}$へ変換する．
  合計ワイヤ本数を$m$，$i\in\{1, ..., m\}$として
  CubiX原点からみた$i$番目のCubiXワイヤ端点の位置を$\bm{r}_i$，
  そのCubiXワイヤ端点から環境ワイヤ端点への単位ベクトルを$\bm{s}_i$とする．
  このとき，$6\times m$行列のワイヤ行列$\bm{W}(\bm{q})$を用いると，
  張力$\bm{f}$によって発揮されるレンチ$\bm{w}$が\equref{eq:wire_matrix}のように表される．
  ただし，張力$\bm{f}$は$i$番目のワイヤ張力を$f_i$として，$\bm{f} = [f_1, ..., f_m]^\top$であり，引張り方向が正である．
  \begin{equation}
    \begin{split}
      \bm{w} &= {\bm{W}(\bm{q})}{\bm{f}},\;\; \mathrm{where}\;\;
      \bm{W}(\bm{q}) =
      \left[ \begin{array}{ccc}
        \bm{s}_{1} & \cdots & \bm{s}_{m} \\
        \bm{r}_{1}\times\bm{s}_{1} & \cdots & \bm{r}_{m}\times\bm{s}_{m}
      \end{array} \right]
    \end{split}
    \label{eq:wire_matrix}
  \end{equation}
  これに基づき，筋骨格系ヒューマノイドの制御のために研究された張力に基づく関節空間制御器\cite{7803367}を参考に，
  \equref{eq:quad_prog}の二次計画問題を解くことで，目標レンチ$\bm{w}^\mathrm{ref}$を発揮する張力$\bm{f}^\mathrm{ref}$を決定する．
  ただし，$\bm{\Lambda}$は重み行列であり，$\bm{f}^\mathrm{min},\bm{f}^\mathrm{max}$はワイヤ張力の最小，最大値ベクトルである．
  \begin{equation}
    \begin{aligned}
      & \underset{\bm{f}^\mathrm{ref}}{\text{min}}
      & & \|\bm{f}^\mathrm{ref}\|^2 + \left[\bm{w}^\mathrm{ref} - {\bm{W}(\bm{q})}{\bm{f}^\mathrm{ref}}\right]^\top \bm{\Lambda} \left[\bm{w}^\mathrm{ref} - {\bm{W}(\bm{q})}{\bm{f}^\mathrm{ref}}\right]\\
      & \text{s.t.}
      & & \bm{f}^\mathrm{min} \leq \bm{f}^\mathrm{ref} \leq \bm{f}^\mathrm{max} \\
    \end{aligned}
    \label{eq:quad_prog}
  \end{equation}
  ここで，巻取りプーリ慣性トルクを補償する張力と，軸摩擦損失トルクを補償する張力とを計算し，
  $\bm{f}^{\mathrm{ref}}$に加えることで$\bm{f}^\mathrm{ref}_{\mathrm{final}}$を得る．
  なお，これらの計算には加速度$\bm{\ddot{q}}^\mathrm{ref}$とワイヤ速度$\bm{\dot{l}}$を用いる．
  % ワイヤ速度$\bm{\dot{l}}$の符号を用いて計算される軸摩擦損失トルクと，
  % $\bm{\ddot{q}}^{ref}$から計算される巻取りプーリ慣性トルクを補償する張力を
  % $\bm{f}^{\mathrm{ref}}$に加えることで$\bm{f}^\mathrm{ref}_{\mathrm{final}}$を得る．

  最後に，\figref{fig:position_control_flow}中のCurrent Calculatorで
  \equref{eq:current}に示す通り，$\bm{f}^\mathrm{ref}_{\mathrm{final}}$を力学モデルに基づいて定数倍することにより，
  各モータに司令する電流$\bm{i}$を得る．
  ただし，$r$は巻取りプーリ半径，$\mu_p$はプーリ伝達効率，$\mu_G$はギアヘッド伝達効率，$G$は減速比，$K_t$はトルク定数である．
  \begin{equation}
    \bm{i} = \cfrac{r}{\mu_p \mu_G G K_t}\bm{f}^\mathrm{ref}_\mathrm{final}
    \label{eq:current}
  \end{equation}

  以上より，CubiXを経路に追従させるためのワイヤ張力計算による位置制御が達成される．
}%

\subsection{Wire Deployment with Flying Anchor} \label{subsec:control flying anchor}
\switchlanguage%
{%

  The wire deployment control using the Flying Anchor is shown in \figref{fig:flying_anchor_placement}.
  In this setup, wires are arranged on pillars fixed to the ceiling, each measuring 350 mm $\times$ 700 mm.
  The relative pose of the pillars is estimated using the AprilTag\cite{Wang2016} attached to the pillars and
  visual odometry executed internally by the drone. 
  By following specified paths, the wires are fastened to the pillars.
  The drone communicates with the PC mounted on CubiX via Wi-Fi,
  through which it transmits images and visual odometry results to the PC and receives target velocity vectors as control commands.
  On CubiX's PC, 
  image-based pose estimation using the AprilTag and PID control of the drone for tracking the target pose are performed.

}%
{%
  % ドローンによるワイヤ配置
  \begin{figure}[tbp]
    \begin{center}
      \includegraphics[width=1.0\columnwidth]{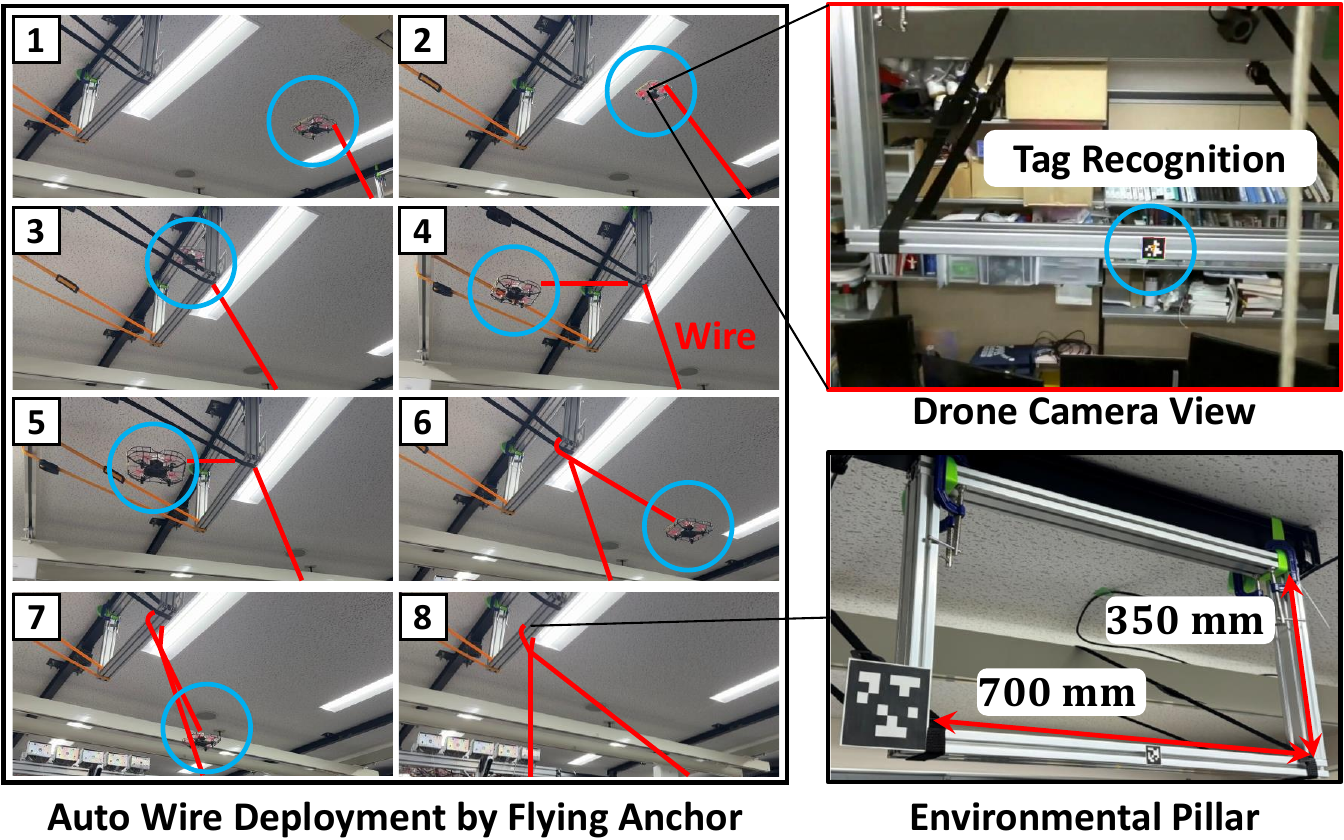} 
      \vspace{-3.0ex}
      \caption{System of wire deployment using the flying anchor.
      The relative pose of the flying anchor and the pillar is estimated using tag recognition, 
      and the wire is connected to the pillar following the given path.} 
      \vspace{-3.0ex}
      \label{fig:flying_anchor_placement}
    \end{center}
  \end{figure}

  % 図を書いたら　図を見ながら
  % AR+内部MVOで自己位置　外部のPC　PtoP制御　
  Flying Anchorを用いたワイヤ配置制御を\figref{fig:flying_anchor_placement}に示す．
  ここでは，実験環境として天井に固定された寸法350 mm $\times$ 700 mmの柱にワイヤを配置した．
  柱につけたARマーカと，ドローンが内部で実行するビジュアルオドメトリを用いて柱相対の自己位置を推定し，
  指定された経路をたどることでワイヤを柱に結びつける．
  ドローンはCubiXに搭載したPCにWi-Fiを介して接続されており，そのPCに画像とビジュアルオドメトリ結果を送信し，
  制御指令値として目標速度ベクトルを受信して動作する． % RCコマンドってなんていえば良いんだ
  CubiXのPCでは，AprilTag\cite{Wang2016}を用いた画像による自己位置推定と目標位置に追従するためのPID制御が行われている．
}%

\section{Experiments} \label{sec:experiments}
\subsection{Spatial Movement Driven by 8 Wires}
\switchlanguage%
{%
  In this experiment, CubiX is driven by the maximum number of wires it can wind, which is 8 wires,
  to perform spatial movements.
  The aim is to evaluate whether the design and control of CubiX function correctly as a wire-driven robot.
  A frame measuring 1 m $\times$ 1 m $\times$ 1 m is prepared as the environment, 
  with 8 wires evenly distributed in all directions.
  CubiX's operation and wire placement are shown in \figref{fig:exp1_pic}, 
  and the time-series data of $\bm{q},\; \bm{q}^\mathrm{ref}$ are shown in \figref{fig:exp1_data}.
  Using the hook attached to CubiX, it lifted a 1.0 kg box to a platform 0.45 m above.

  From \figref{fig:exp1_pic}, 
  it is evident that CubiX successfully lifted and placed the box onto the platform during spatial movement.
  Furthermore, from \figref{fig:exp1_data},
  it can be observed that CubiX tracked the target position and orientation without exhibiting significant oscillations,
  except for the portions indicated by the red circle.
  This experiment demonstrated that CubiX, equipped with a total of 8 wire modules and connected wires to the environment,
  can follow its intended path by calculating and winding the necessary tensions of these wires. 

  However, the error from the target occurred during the specific interval indicated by the red circle. 
  Observing the corresponding part marked with a green circle in the graph of $\bm{f}^\mathrm{ref}_\mathrm{final}$ in \figref{fig:exp1_data}, 
  it can be seen that 2 wires hit their $\bm{f}^\mathrm{max}$. 
  This indicates that the Jacobian matrix $\bm{W}(\bm{q})$ at this pose of CubiX is unable to represent the target wrench.
  When the feasible wrench is biased in one direction, the optimization in \equref{eq:quad_prog} may fail to solve, 
  leading to instability if feedback wrenches are demanded in directions difficult to exert. 
  To address this issue, adjusting the optimization weights $\bm{\Lambda}$ or designing the wire placement 
  with large wrenches in all directions could be considered.

  \begin{figure}[tbp]
    \begin{center}
      \includegraphics[width=1.0\columnwidth]{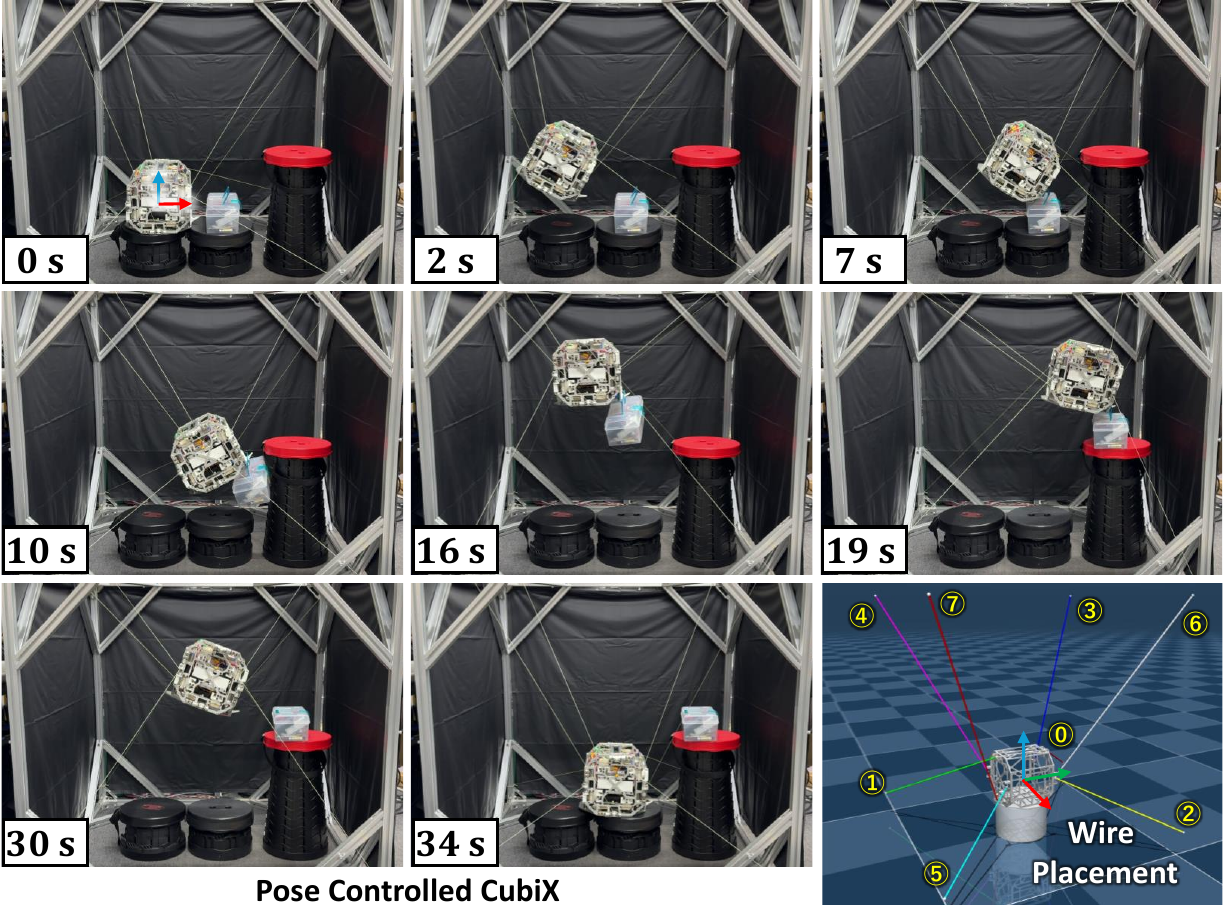}
      \vspace{-4.5ex}
      \caption{Pose controlled CubiX and the wire placement.
      CubiX correctly followed the given path and transported the box onto the platform.}
      \vspace{-3.0ex}
      \label{fig:exp1_pic}
    \end{center}
  \end{figure}
  \begin{figure}[tbp]
    \begin{center}
      \includegraphics[width=1.0\columnwidth]{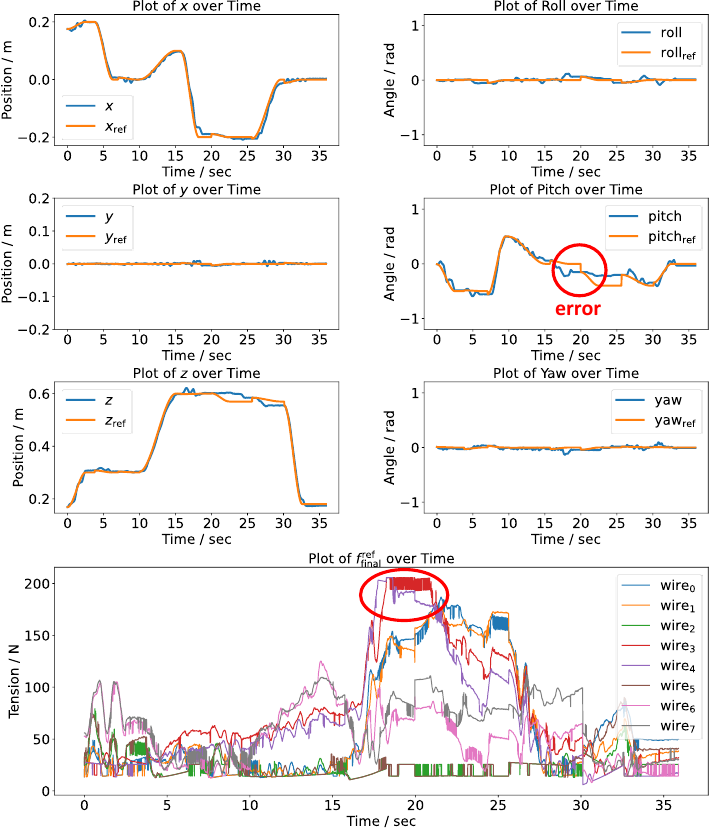}
      \vspace{-4.5ex}
      \caption{$\bm{q},\;\bm{q}^\mathrm{ref}$ and $\bm{f}^\mathrm{ref}_\mathrm{final}$ in 8-wire drive experiment.
      These results show that the current pose of CubiX followed the target pose, 
      except for the area indicated by the red circle.
      }
      \vspace{-5.5ex}
      \label{fig:exp1_data}
    \end{center}
  \end{figure}

  \begin{figure}[t]
    \begin{center}
      \includegraphics[width=1.0\columnwidth]{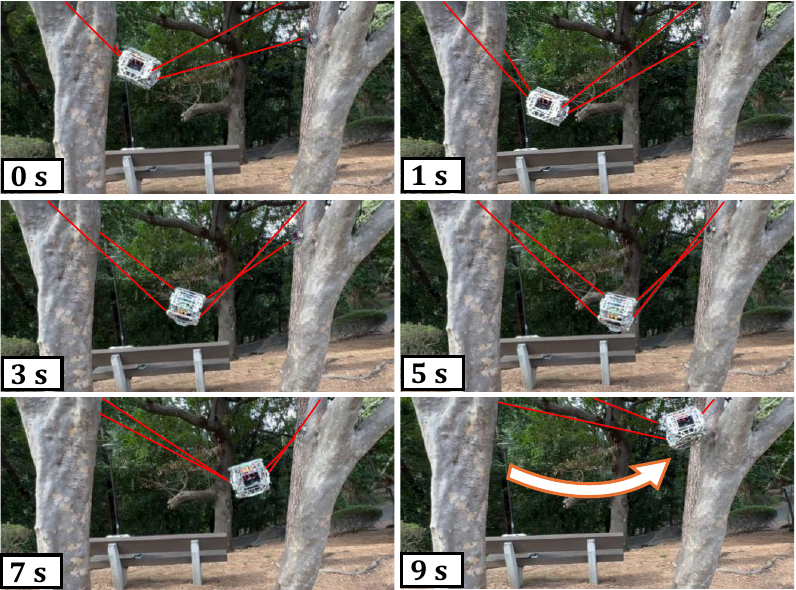}
      \vspace{-4.5ex}
      \caption{CubiX in outdoor experiment.
      CubiX is connected to the environment by 4 wires, 2 on each side of the tree.
      It shows that CubiX has moved from the tree on the left to the tree on the right.}
      \vspace{-3.0ex}
      \label{fig:exp2_pic}
    \end{center}
  \end{figure}
  \begin{figure}[t]
    \begin{center}
      \includegraphics[width=1.0\columnwidth]{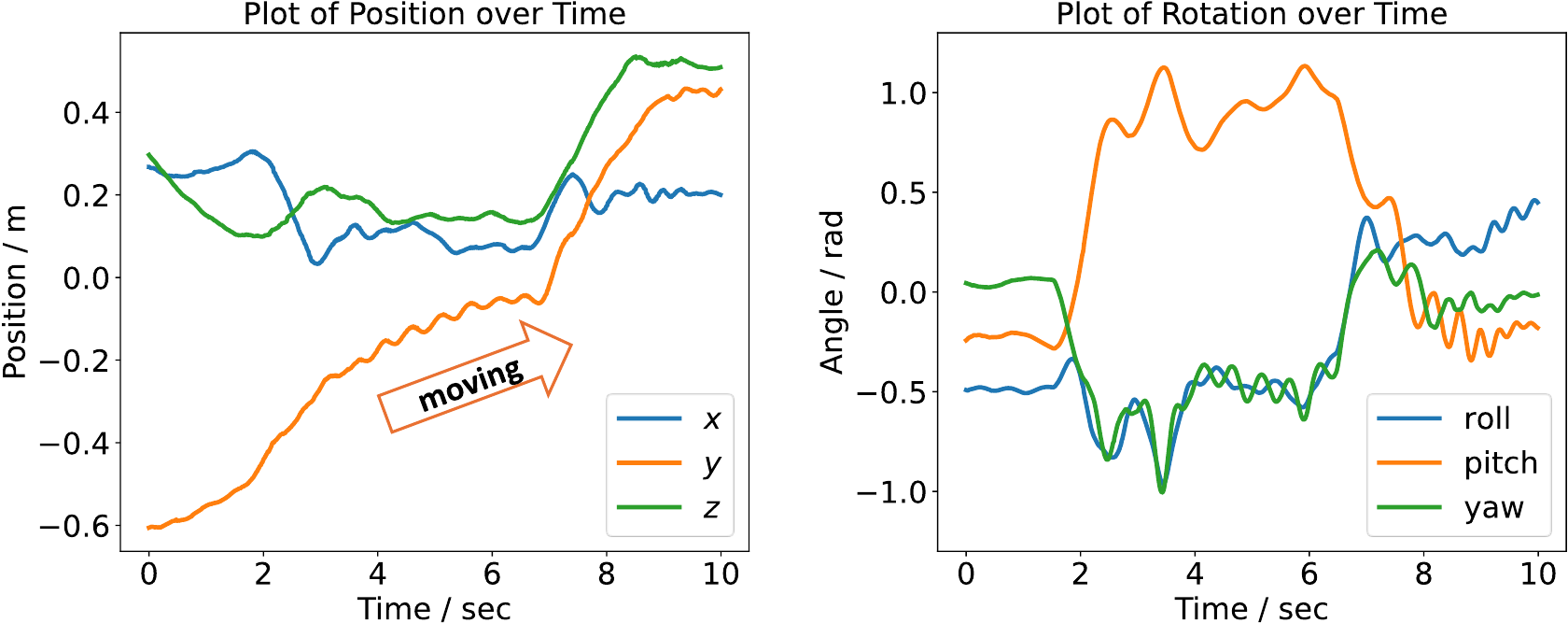}
      \vspace{-4.5ex}
      % \caption{$\bm{q},\;\bm{q}^\mathrm{ref}$ in outdoor experiment}
      \caption{$\bm{q}$ in outdoor experiment.
      The increase in the y direction from -0.5 m to 0.5 m indicates that CubiX is moving from tree to tree.}
      \vspace{-4.0ex}
      \label{fig:exp2_data}
    \end{center}
  \end{figure}

  \begin{figure}[t]
    \begin{center}
      \includegraphics[width=1.0\columnwidth]{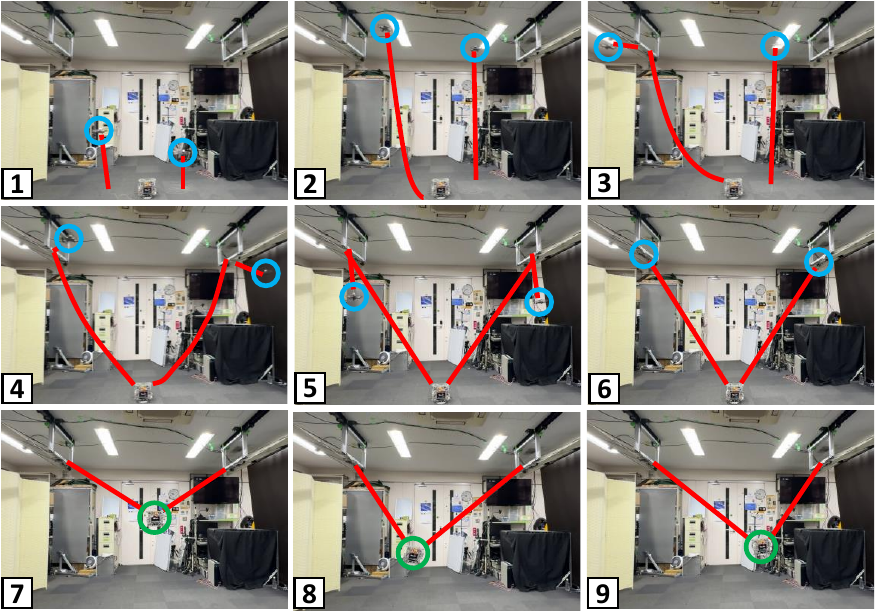}
      \vspace{-4.5ex}
      \caption{CubiX and 2 flying anchors in wire deployment experiment.
      It shows that wires were connected to pillars in the environment by flying anchors
      and that CubiX was able to be driven by these wires.}
      \vspace{-3.0ex}
      \label{fig:exp3_pic}
    \end{center}
  \end{figure}
  \begin{figure}[t]
    \begin{center}
      \includegraphics[width=1.0\columnwidth]{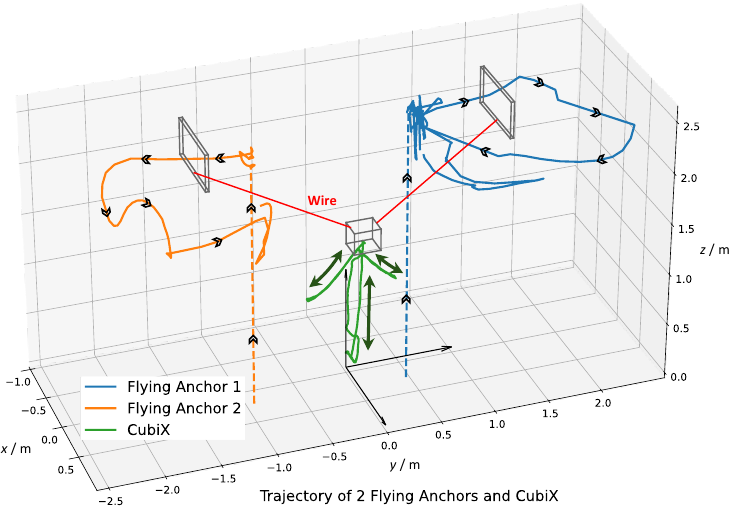}
      \vspace{-3.0ex}
      % \caption{$\bm{q},\;\bm{q}^\mathrm{ref}$ in outdoor experiment}
      \caption{Trajectory of 2 flying anchors and CubiX.
      Until the flying anchor recognizing the AprilTag, its trajectory is indicated by dashed lines.
      It shows that the wires were connected to the pillars by 2 flying anchors circling around the pillars. 
      It also shows that CubiX used those wires to move up and down and left and right.}
      \vspace{-4.0ex}
      \label{fig:exp3_data}
    \end{center}
  \end{figure}
}%
{%
  \begin{figure}[tbp]
    \begin{center}
      \includegraphics[width=1.0\columnwidth]{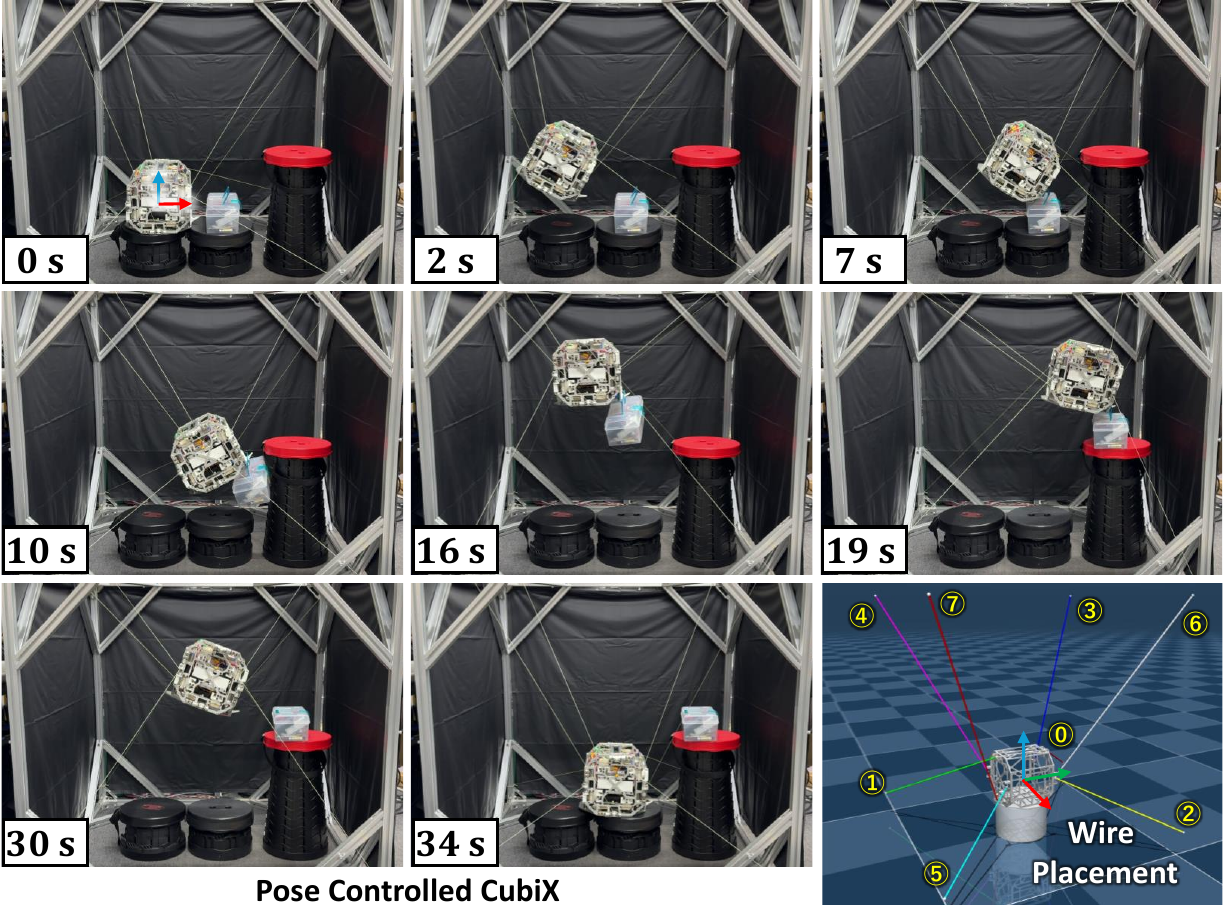}
      \vspace{-3.0ex}
      \caption{Pose controlled CubiX and the wire placement.
      CubiX correctly followed the given path and transported the box onto the platform.}
      \vspace{-3.0ex}
      \label{fig:exp1_pic}
    \end{center}
  \end{figure}
  \begin{figure}[tbp]
    \begin{center}
      \includegraphics[width=1.0\columnwidth]{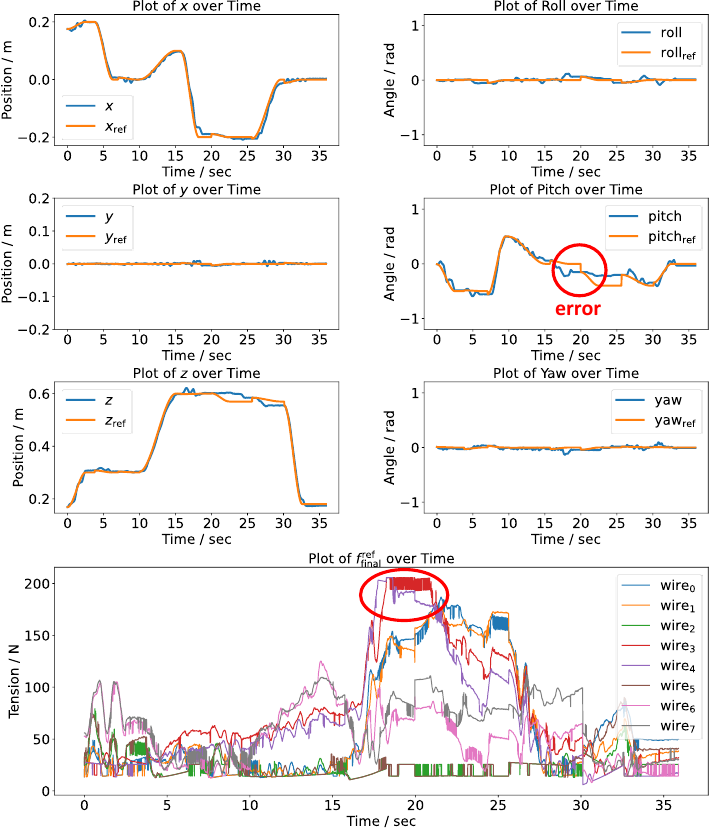}
      \vspace{-3.0ex}
      \caption{$\bm{q},\;\bm{q}^\mathrm{ref}$ in 8-wire drive experiment.
      These results show that the current pose of CubiX followed the target pose, 
      except for the area indicated by the red arrow.
      }
      \vspace{-3.0ex}
      \label{fig:exp1_data}
    \end{center}
  \end{figure}
  この実験では，CubiXが巻き取れる最大のワイヤ本数である8本のワイヤで駆動し，空間移動を行うことで，
  CubiXにおける設計と制御がワイヤ駆動ロボットとして正しく機能するかを評価する．
  環境として寸法1 m $\times$ 1 m $\times$ 1 mのアルミフレームを用意し，全方位に均等に8本のワイヤを配置した．
  CubiXが動作する様子を\figref{fig:exp1_pic}に，
  $\bm{q},\; \bm{q}^\mathrm{ref}$の時系列データを\figref{fig:exp1_data}に示す．
  % CubiXを台に乗せ，地面から浮いた状態から実験をはじめ，
  % $z$方向の目標位置を0.5mにした後，$x,y$方向の目標位置を共に-0.1mにしてから0.1mにして空間移動を行った．
  CubiXに取り付けたフックを用いて，
  重さ$1.0 \mathrm{kg}$の箱を$0.45 \mathrm{m}$上の台へ持ち上げる空間移動を行った．
  \begin{figure}[t]
    \begin{center}
      \includegraphics[width=1.0\columnwidth]{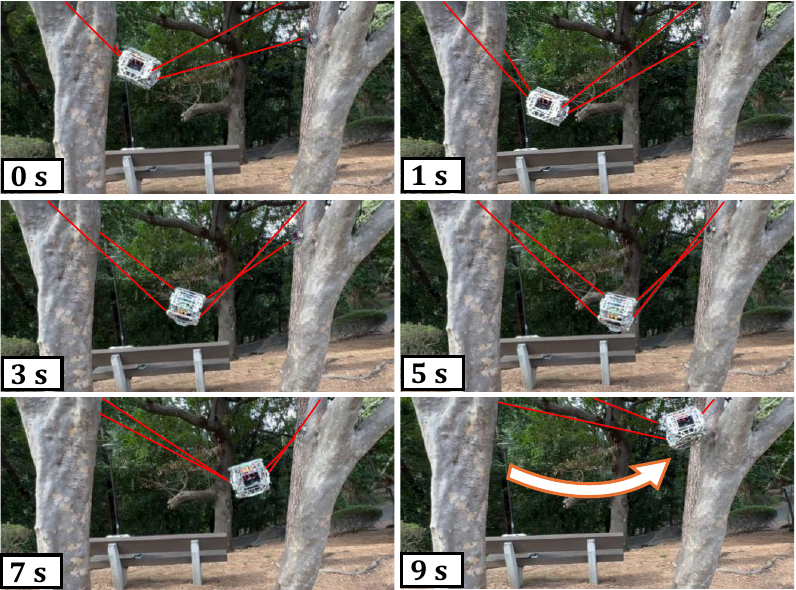}
      \vspace{-3.0ex}
      \caption{CubiX in outdoor experiment.
      CubiX is connected to the environment by four wires, two on each side of the tree.
      It shows that CubiX has moved from the tree on the left to the tree on the right.}
      \vspace{-3.0ex}
      \label{fig:exp2_pic}
    \end{center}
  \end{figure}
  \begin{figure}[t]
    \begin{center}
      \includegraphics[width=1.0\columnwidth]{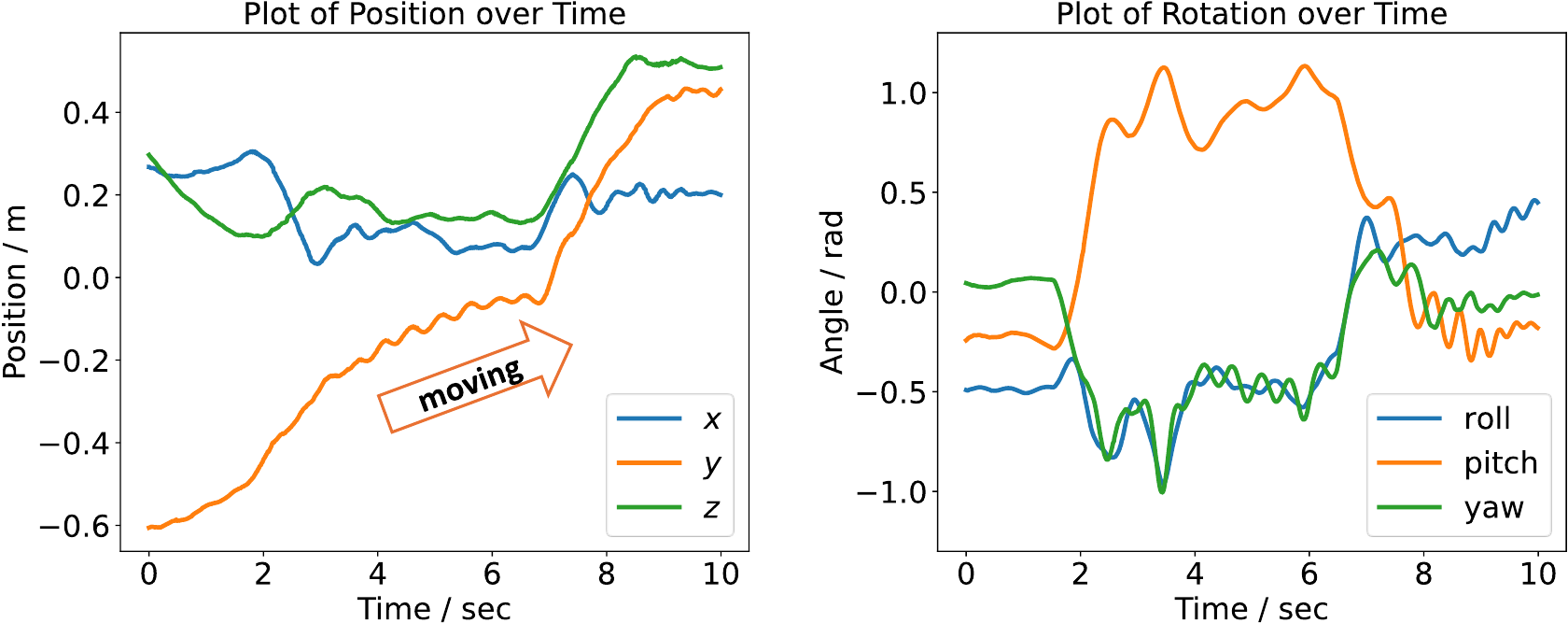}
      \vspace{-3.0ex}
      % \caption{$\bm{q},\;\bm{q}^\mathrm{ref}$ in outdoor experiment}
      \caption{$\bm{q}$ in outdoor experiment.
      The increase in the y direction from -0.5 m to 0.5 m indicates that CubiX is moving from tree to tree.}
      \vspace{-3.0ex}
      \label{fig:exp2_data}
    \end{center}
  \end{figure}

  % 結果
  \figref{fig:exp1_pic}から，CubiXが空間移動することにより，箱を台へ持ち上げて
  置くという操作ができていることがわかる．
  また，\figref{fig:exp1_data}より，pitchの時系列グラフに赤矢印で指し示した部分を除いて，
  CubiXが目標位置，姿勢に振動を起こさずに追従していることがわかる．
  % 考察
  これより，計8個のワイヤモジュールを搭載したCubiXが，体外にワイヤを配置し，
  それらの必要張力を計算して巻き取ることで，目標経路に自身を追従させる能力を持つことが示された．
  一方，赤矢印で指し示した部分で目標に追従できていない原因は，
  そのワイヤ配置と位置，姿勢におけるワイヤ行列$\bm{W}(\bm{q})$では
  いかなる張力を出しても目標レンチ$\bm{w}^{\mathrm{ref}}$を表現できないことにある．
  発揮可能なレンチがある方向に偏っていると，
  発揮しづらい方向にフィードバックレンチが要求された場合に\equref{eq:quad_prog}の最適化が解けなくなり，挙動が不安定になる．
  これは，最適化の重み$\bm{\Lambda}$を調節するか，
  発揮可能レンチがあらゆる方向に大きいワイヤ配置を設計することで改善されると考えられる．
  % 空間移動達成
  % ワイヤ駆動パラレルロボットや
  % 赤の部分はワイヤ配置で発揮可能レンチがないよ
}%

\subsection{Outdoor Experiment}
\switchlanguage%
{%
  
  In this experiment, CubiX was taken outdoors to demonstrate wire-driven movement by connecting wires to trees,
  showcasing CubiX as a portable CDPR. 
  Two wires were placed on each of the trees on either side of CubiX, totaling 4 wires, to perform spatial movement.
  The operation of CubiX is shown in \figref{fig:exp2_pic}, and the time-series data of $\bm{q}$ is shown in \figref{fig:exp2_data}. 
  CubiX was driven by commanding target tensions directly to each wire.

  In \figref{fig:exp2_data}, focusing on the $y$-direction, it is observed to increase from -0.5 m to 0.5 m,
  indicating CubiX's movement from the left tree to the right tree.
  The fact that CubiX was able to connect itself to the environment via wires and
  drive by winding them up in outdoor location
  demonstrates that the developed CubiX is a portable CDPR integrated with the necessary devices
  such as actuators, PC, battery, and sensor for operation.
  Furthermore, CubiX in this experiment is driven by 4 wires, 
  resulting in an underactuated system since CubiX has six degrees of freedom.
  However, it successfully accomplishes the movement from the left tree to the right one.
  This indicates that by altering the wire placement or the number of wires,
  parallel wire-driven structures can be formed to match the environment in which CubiX connects and uses.
}%
{%
  \begin{figure}[t]
    \begin{center}
      \includegraphics[width=1.0\columnwidth]{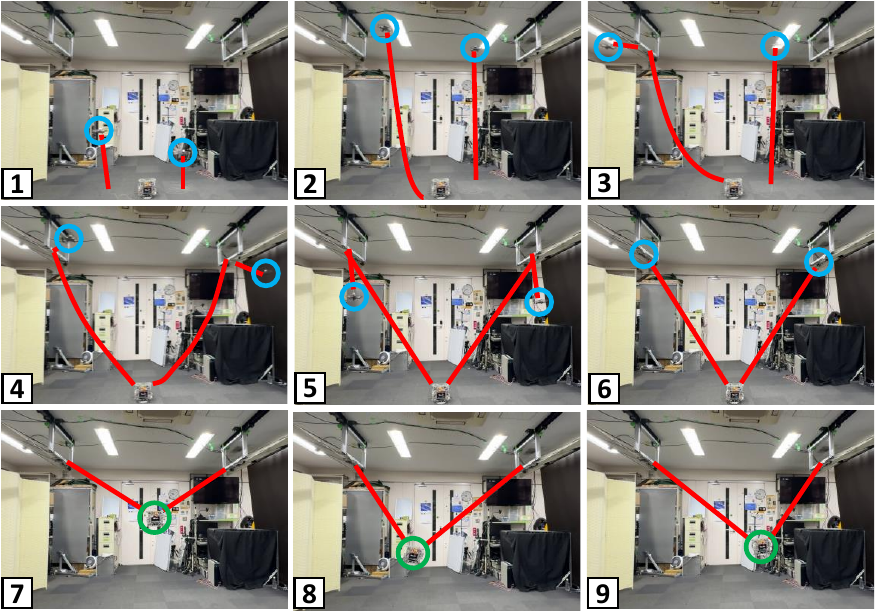}
      \vspace{-3.0ex}
      \caption{CubiX and two flying anchors in wire deployment experiment.
      It shows that wires were connected to pillars in the environment by flying anchors
      and that CubiX was able to be driven by these wires.}
      \vspace{-3.0ex}
      \label{fig:exp3_pic}
    \end{center}
  \end{figure}
  \begin{figure}[t]
    \begin{center}
      \includegraphics[width=1.0\columnwidth]{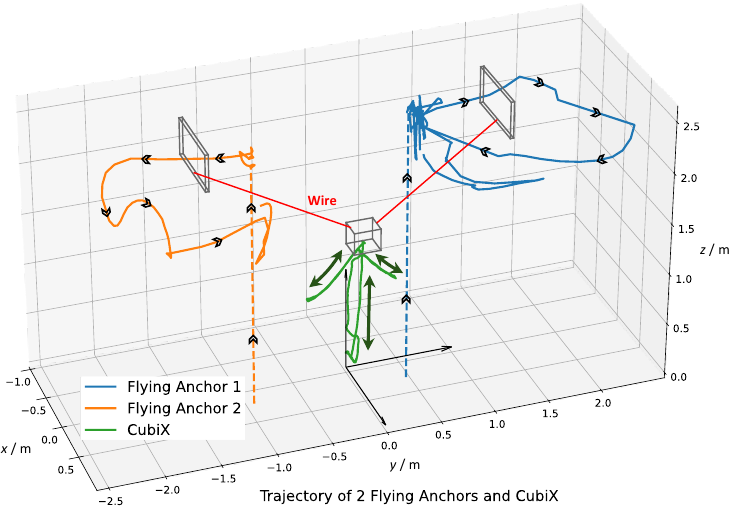}
      \vspace{-3.0ex}
      % \caption{$\bm{q},\;\bm{q}^\mathrm{ref}$ in outdoor experiment}
      \caption{Trajectory of 2 flying anchors and CubiX.
      Until the Flying Anchor recognizing the AprilTag, its trajectory is indicated by dashed lines.
      It shows that the wires were connected to the pillars by two flying anchors circling around the pillars. 
      It also shows that CubiX used those wires to move up and down and left and right.}
      \vspace{-3.0ex}
      \label{fig:exp3_data}
    \end{center}
  \end{figure}
  この実験では，CubiXを屋外に持ち出し，ワイヤ端点を木に接続してワイヤ駆動を実現することで，
  CubiXがポータブルなCDPRであるということを示す．
  CubiXの左右にある木にそれぞれ2本ずつワイヤを配置して，計4本のワイヤによって空間移動を行った．
  CubiXが動作する様子を\figref{fig:exp2_pic}に，
  % $\bm{q},\; \bm{q}^\mathrm{ref}$の時系列データを\figref{fig:exp2_data}に示す．
  $\bm{q}$の時系列データを\figref{fig:exp2_data}に示す．
  各ワイヤに直接，目標張力を司令することでCubiXを駆動させた．

  % 結果
  \figref{fig:exp2_data}において，$y$方向に注目すると，-0.5mから0.5mへ増加していることがわかり，
  CubiXが左の木から右の木へ移動していることがわかる．
  % more TODO
  % また，
  % 考察
  CubiXが自身をワイヤによって環境に接続し，それを自ら巻き取ることによって駆動できたことから，
  開発したCubiXはアクチュエータやPC，バッテリ，センサなどの
  稼働に必要なデバイスを内蔵したポータブルなCDPRであることが示された．
  なお，この実験は4本のワイヤによる駆動になっており，
  CubiXの自由度数は6であるため劣駆動となっているが，
  左の木から右の木へ移るという移動を果たしている．
  つまり，ワイヤ配置やワイヤ本数を変更することで，
  接続して利用する環境
  % や目的
  に合わせたパラレルワイヤが形成できることもわかる．

  % なお，劣駆動となっているが，同様に$\Lambda$を...
  % ワイヤ配置がちゃっかり可変
  % 環境に合わせて
  % CubiXがポータブルなCDPRとして活動場所でパラレルワイヤ駆動を形成できることが確認できた．
}%

\subsection{Wire Deployment by Flying Anchors}
\switchlanguage%
{%
  In this experiment, 2 flying anchors were used to connect CubiX's wires to the environment,
  followed by CubiX's spatial movement,
  demonstrating CubiX's ability to autonomously connect wires to the environment and drive itself using them.
  Following the method outlined in \subsecref{subsec:control flying anchor},
  CubiX was driven by directly commanding tension after connecting the wires to 2 pillars on the ceiling using the flying anchors.
  The operation of the 2 flying anchors and CubiX is shown in \figref{fig:exp3_pic},
  while their trajectories are shown in \figref{fig:exp3_data}.

  From \figref{fig:exp3_data}, it is evident that the 2 flying anchors maneuver through the pillars,
  adjust their pose using AprilTags, and connect the wires to the pillars by circumnavigating them.
  Additionally, it can be observed that CubiX moves vertically and horizontally using these wires.
  This outcome demonstrates that CubiX, a portable wire-driven parallel robot,
  can autonomously connect wires to the environment using flying anchors and drive itself using them.
}%
{%
  この実験では，
  2台の飛行アンカーによりCubiXのワイヤを環境に接続し，その後にCubiXが空間移動することで，
  CubiXが自律的にワイヤを環境へ接続し，それを用いて駆動できることを示す．
  \subsecref{subsec:control flying anchor}で示した手法により，
  飛行アンカーを用いて
  %自動で
  ワイヤを天井に固定した2つの柱に結びつけた後，張力を直接司令することによりCubiXを駆動した．
  2台の飛行アンカーとCubiXが動作する様子を\figref{fig:exp3_pic}に，
  それらの軌跡を\figref{fig:exp3_data}に示す．

  % 結果
  \figref{fig:exp3_data}より，
  2台の飛行アンカーがARタグで目標位置を微調節しながら柱をくぐり抜け，
  柱を1周することでワイヤを柱に接続していることがわかる．
  また，CubiXがそのワイヤを用いて上下，左右に空間を移動していることもわかる．
  % 考察
  この結果より，ポータブルなワイヤ駆動パラレルロボットであるCubiXが
  飛行アンカーを用いてワイヤを自ら環境へ接続し，
  それを用いて駆動できることが示された．
}%

% \section{Discussion} \label{sec:discussion} % ２つの実験の毛色が違うからそれぞれで考察したらいいのかな
% \switchlanguage%
% {%
%   hoge
% }%
% {%
%   8本ワイヤ駆動の実験について，hogeに注目する．
%   % 実験1より，CubiXはワイヤ駆動ロボットとして，十分な制御性能を持っていることがわかることを述べる
%   % できていない部分はワイヤ配置によることを示す
% 
%   屋外での実験について，hogeに注目すると，
%   % 実験2より，CubiXは屋外において環境にワイヤを接続することでワイヤ駆動ロボットとなり，ポータブルな
%   % ワイヤ駆動ロボットであることを述べる
% }%

\section{Conclusion} \label{sec:conclusion}
\switchlanguage%
{%
  In this study, we developed CubiX, a wire-driven robot connecting to and utilizing the environment,
  and conducted experiments involving spatial movement with 8-wire drive, outdoor parallel wire-drive experiments,
  and wire deployment experiments using flying anchors.
  CubiX, a cube-shaped wire-driven robot,
  is equipped with 8 wire modules capable of continuous tensioning of up to 180 N each,
  enabling it to autonomously control its pose through wire tension calculations.
  The spatial movement experiments with 8-wire drive demonstrated CubiX's ability to 
  control its pose by connecting with the environment through wires and using them for driving.
  The outdoor parallel wire-drive experiments confirmed CubiX's capability to form parallel wire drives
  as a portable CDPR in outdoor locations.
  Moreover, the wire deployment experiments with flying anchors showcased CubiX's autonomous ability to connect wires
  to the environment and utilize them for movement.

  As future works,
  optimizing wire placement tailored to specific tasks and integrating tools such as robotic arms 
  and carts with CubiX could expand its capabilities,
  enabling CubiX to utilize the environment and exhibit performance unconstrained by its physical structure.
}%
{%
  本研究では，環境に接続するワイヤ駆動ロボットCubiXを開発し，
  8本のワイヤ駆動による空間移動実験と屋外におけるパラレルワイヤ駆動実験，
  飛行アンカーによるワイヤ配置実験を行った．
  CubiXは最大連続張力180 N のワイヤ巻取りモジュールを計8個体内に搭載した
  キューブ型のワイヤ駆動ロボットであり，
  ワイヤ張力計算によって自身の位置を自律的に制御することが可能である．
  8本のワイヤ駆動による空間移動実験から，
  CubiXが自身と環境とをワイヤで接続し，それを巻き取って駆動することで位置制御ができることが確認でき，
  屋外におけるパラレルワイヤ駆動実験から，
  CubiXがポータブルなCDPRとして活動場所でパラレルワイヤ駆動を形成できることが確認できた．
  また，飛行アンカーによるワイヤ配置実験から，
  CubiXが自律的にワイヤを環境に接続し，それを用いて駆動できることが確認できた．

  今後の展望として，
  必要なタスクに応じてワイヤ配置を最適化することや，
  CubiXにロボットアームや台車などの道具を合体させて能力を拡張することが実現すれば，
  CubiXは環境を巧みに利用し身体構造にとらわれない性能を発揮するロボットとなることが考えられる．
}%

{
  %\footnotesize
  %\small
  %\bibliographystyle{junsrt}
  \bibliographystyle{IEEEtran}
  \bibliography{bib}
}

\end{document}